%% file: main.tex
\documentclass[10pt,journal,compsoc]{IEEEtran}
\usepackage{hyperref}       
\usepackage{url}            
\usepackage{booktabs}       
\usepackage{amsfonts}       
\usepackage{nicefrac}       
\usepackage{microtype}      
\usepackage{enumerate}
\usepackage{graphicx}
\usepackage{float}
\usepackage{subfigure}
\usepackage{booktabs}
\usepackage{adjustbox}
\usepackage{multirow}
\usepackage{amsmath}
\usepackage{amssymb}
\usepackage{graphicx}
\usepackage{subfigure}
\usepackage{wrapfig}
\usepackage{enumitem}
\usepackage{hyperref}
\usepackage{pifont}
\usepackage{color}
\usepackage[ruled]{algorithm2e}
\usepackage{algorithmic}
\usepackage{amsmath}
\usepackage{amsthm}
\usepackage{bbm}
\usepackage{caption}
\usepackage[switch]{lineno}

\usepackage{setspace}

\usepackage{ragged2e}
\usepackage{tikz}

\usepackage{array}
\newcolumntype{L}[1]{>{\raggedright\let\newline\\\arraybackslash\hspace{0pt}}m{#1}}
\usepackage{pdfcomment}
\usepackage{marginnote}

\usepackage{mathtools}
\usepackage{paralist}
\let\itemize\compactitem
\let\enditemize\endcompactitem
\let\enumerate\compactenum
\let\endenumerate\endcompactenum

\ifCLASSOPTIONcompsoc
  \usepackage[nocompress]{cite}
\else
  \usepackage{cite}
\fi
\ifCLASSINFOpdf
\else
\fi

\begin{document}
\title{Multimodal Fusion on Low-quality  Data: \\A Comprehensive Survey}

\author{Qingyang Zhang, Yake Wei, Zongbo Han, Huazhu Fu, Xi Peng, Cheng Deng,\\Qinghua Hu, Cai Xu, Jie Wen, Di Hu, Changqing Zhang
\IEEEcompsocitemizethanks{\IEEEcompsocthanksitem Qingyang Zhang, Zongbo Han, Qinghua Hu and Changqing Zhang were with the the College of Intelligence and Computing, Tianjin University (e-mail: {qingyangzhang,
zongbo, zhangchangqing}@tju.edu.cn).
\IEEEcompsocthanksitem Huazhu Fu was with Institute of High Performance Computing (IHPC), Agency for Science, Technology and Research (A*STAR) (e-mail: hzfu@ieee.org).
\IEEEcompsocthanksitem Yake Wei and Di Hu were with Gaoling School of Artificial Intelligence, Renmin University of China (e-mail: {yakewei, dihu}@ruc.edu.cn).
\IEEEcompsocthanksitem Xi Peng was with College of Computer Science, Sichuan University (e-mail: pengxi@scu.edu.cn).
\IEEEcompsocthanksitem Cai Xu was with School of Computer Science and Technology, Xidian University (e-mail: cxu@xidian.edu.cn).
\IEEEcompsocthanksitem Cheng Deng was with School of Electronic Engineering Xidian University (e-mail: chdeng@mail.xidian.edu.cn).
\IEEEcompsocthanksitem Jie Wen is with School of Computer Science and Technology, Harbin Institute of Technology, Shenzhen (e-mail: jiewen pr@126.com).
\IEEEcompsocthanksitem Correspondence to Changqing~Zhang $<$zhangchangqing@tju.edu.cn$>$.
}}

\markboth{Submitted to IEEE Transactions on Pattern Analysis and Machine Intelligence}
{Shell \MakeLowercase{\textit{et al.}}: Bare Demo of IEEEtran.cls for Computer Society Journals}
\IEEEtitleabstractindextext{%
\begin{abstract}
\justifying
Multimodal fusion focuses on integrating information from multiple modalities with the goal of more accurate prediction, which has achieved remarkable progress in a wide range of scenarios, including autonomous driving and medical diagnosis. However, the reliability of multimodal fusion remains largely unexplored especially under low-quality data settings. This paper surveys the common challenges and recent advances of multimodal fusion in the wild and presents them in a comprehensive taxonomy. From a data-centric view, we identify four main challenges that are faced by multimodal fusion on low-quality data, namely (1) noisy multimodal data that are contaminated with heterogeneous noises, (2) incomplete multimodal data that some modalities are missing, (3) imbalanced multimodal data that the qualities or properties of different modalities are significantly different and (4) quality-varying multimodal data that the quality of each modality dynamically changes with respect to different samples. This new taxonomy will enable researchers to understand the state of the field and identify several potential directions. We also provide discussion for the open problems in this field together with interesting future research directions.
\end{abstract}
}
\maketitle
\IEEEdisplaynontitleabstractindextext

\IEEEpeerreviewmaketitle
\input{introduction}

\input{noise/noise_ver2}

\input{incomplete/incomplete}

\input{balance/balanced_mm_ver2}

\input{dynamic/dynamic}

\section{Conclusion}
In this survey, we comprehensively investigate the main challenges of multimodal learning in the wild: noisy, partial, unbalanced, and quality-varying multimodal data. Some of them such as noisy multimodal data have been studied for a long time, but more recent interests in representation and translation have led to a large number of new multimodal algorithms and exciting multimodal applications. For example, low-quality multimodal data for foundation models, e.g., CLIP. (2) low-quality multimodal data for other tasks beyond fusion, e.g., alignment (retrieval). (3) Other low-quality forms beyond noise, incomplete, unbalanced, and dynamic (e.g., adversarial). We believe that our survey will help catalog future research papers and better understand the remaining unresolved problems facing multimodal machine learning.


%












\bibliographystyle{IEEEtran}
\bibliography{main}

\end{document}

%% file: introduction.tex
\IEEEraisesectionheading{\section{Introduction}}\label{sec:introduction}

\IEEEPARstart{O}{ur} perception of the world is based on multiple modalities, e.g., touch, sight, hearing, smell and taste. Even when some sensory signals are unreliable, humans can extract useful clues from imperfect multimodal inputs and further piece together the whole scene of happening events~\cite{rideaux2021multisensory}. With the development of sensory technology, we can easily collect diverse forms of data for analysis. To fully release the value of each modality, multimodal fusion emerges as a promising paradigm to obtain precise and reliable predictions by integrating all available cues for downstream analysis tasks, e.g., medical image analysis, autonomous vehicles~\cite{chang2022fast, xiao2020multimodal} and sentiment recognition~\cite{yadav2023deep, niu2016sentiment, xu2019multi}. Intuitively, fusing information from different modalities offers the possibility of exploring cross-modal correlation and gaining better performance. However, there is growing recognition that widely-used AI models are often misled by spurious correlations and biases within low-quality data. In real-world setting, the quality of different modalities usually varies due to unexpected environmental factors or sensor issues. Some recent studies have shown both empirically and theoretically that conventional multimodal fusion may fail on low-quality multimodal data in the wild, e.g., imbalanced~\cite{wang2020makes,peng2022balanced,wu2022characterizing, huang2022modality}, noisy~\cite{xu2022different} or even corrupted~\cite{huang2021learning} multimodal data. To overcome such a limitation and take a step to strong and generalized multimodal learning in the real world, we identify the properties of low-quality multimodal data and focus on some unique challenges for multimodal machine fusion in real-world setting. We also highlight technical advances that may help to make multimodal fusion more reliable and trustworthy in the open environment. In this paper, we identify and explore four core technical challenges surrounding multimodal fusion on low-quality multimodal data. They are summarized as follows:

\begin{itemize}

\item [(1)] \textbf{Noisy multimodal data.} The first fundamental challenge is learning how to mitigate the underlying influence of arbitrary noise in multimodal data. High-dimensional multimodal data tend to contain complex noise. The heterogeneity of multimodal data makes it challenging while also provides chances to identify and reduce the potential noise by exploring the correlation among different modalities.

\item [(2)] \textbf{Incomplete multimodal data.} The second fundamental challenge is learning with incomplete multimodal data. For instance, in the medical field, patients even with the same disease may choose different medical examinations producing incomplete multimodal data. Developing flexible and reliable multimodal learning methods that can handle incomplete multimodal data is a challenging yet promising research direction.

\item [(3)] \textbf{Imbalanced multimodal data.} The third fundamental challenge is how to mitigate the influence of bias and discrepancy between modalities. For example, the visual modality is more effective than the audio modality on the whole, leading the model to take shortcuts and lack exploration of audio. Although existing fusion methods exhibit promising performance, they may fail to perform better than the unimodal predominant model for inference on some modality-preferred applications.

\item [(4)] \textbf{Quality dynamically varying multimodal data.} The fourth fundamental challenge is how to adapt the quality dynamically varying nature of multimodal data. In practice, the quality of one modality often varies for different samples due to unforeseeable environmental factors or sensor issues. For example, the RGB image is less informative than the thermal modality in low-light or backlight conditions. Therefore, dynamically integrating multimodal data by being aware of the varying quality is necessary in real-world applications.
\end{itemize}

Towards addressing these increasingly important issues of multimodal fusion, this study systematically organizes the key challenges through several taxonomies. Different from previous related works discuss about various multimodal learning tasks~\cite{xu2013survey, baltruvsaitis2018multimodal}, this survey is mainly focusing on multimodal fusion which is the most fundamental problem in multimodal learning and the unique challenges raising upon low-quality multimodal data in downstream tasks, including clustering, classification, object detection and semantic segmentation. In the following sections, we detailed this field by recent advances and technical challenges of multimodal fusion facing low-quality data: learning on noisy multimodal data (Section 2), missing modality imputation (Section 3), balanced multimodal fusion (Section 4), and dynamic multimodal fusion (Section 5). A discussion in Section 6 is provided as a conclusion.

\begin{figure*}[!t]
    \centering
    \includegraphics[width=0.99\textwidth]{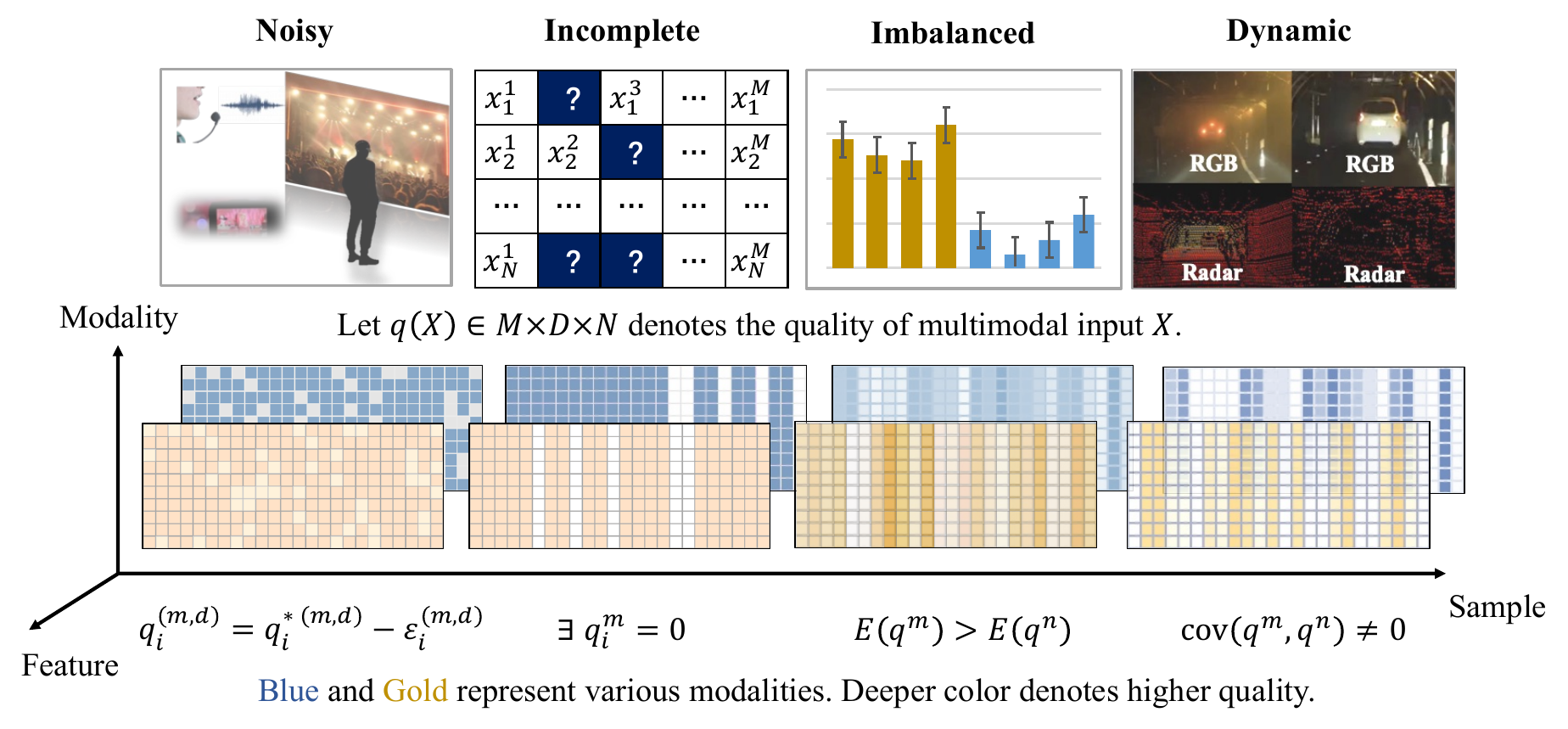}
    \caption{\textbf{Illustrations of challenges for machine learning on low-quality multimodal data. Blue and gold represent various modalities. Deeper color denotes higher quality. Assuming we have $N$ multimodal samples consisting $M$ different modalities, the dimension of each modality is $D$. $q(x)$ denotes the quality of multimodal input, i.e., the information collected from $x$ that can support the downstream tasks. (a) The quality of noisy multimodal data is randomly influenced by unexpected environmental factors. (b) Certain modalities of incomplete multimodal data are of zero quality (do not convey any useful information). (c) The expected quality of modalities are different for imbalanced multimodal data. (d) The quality of different modality are varying for samples. }} 
    \label{fig:cover}
\end{figure*}

%% file: noise/noise_ver2.tex
\section{Learning on Noisy Multimodal Data}
Collecting high-quality multimodal data in real-world scenarios inevitably faces significant challenges due to the presence of noise. Multimodal data \cite{kulkarni2020pixel} noise can arise from sensor errors~\cite{DBLP:conf/cvpr/ChengZDJL19}, environmental interference, or transmission losses. For vision modality, the electronic noise in sensors leads to detail loss. Besides, audio modality may suffer from unexpected distortion due to environmental factors. To make things worse, weakly aligned or even unaligned multimodal samples are also commonly presented, which can be seen as more complex noise that lies in a higher-level semantic space. Fortunately, considering the correlation between multiple modalities or better utilizing multimodal data can help the fusion of noisy multimodal data. Various related works \cite{salvi2023multi,DBLP:conf/cvpr/BaiHZHCFT22,DBLP:conf/cvpr/ChengZDJL19} demonstrate that multimodal models surpass their unimodal counterparts. This can be attributed to the ability of multimodal data to identify and mitigate potential noise by leveraging the correlations among different modalities.

Multimodal noise can be roughly split into two categories according to their source: 1) modal-specific noise that arises from sensor errors, environmental factors, or transmission of each modality separately, 2) cross-modal noise that arises from weakly aligned or unaligned multimodal pairs which can be regarded as semantic level noise.

\subsection{Modal-specific noise reduction} 
Modal-specific noise reduction methods highly depends on the input modalities and task at hand. In this section, we focus on \textbf{visual} noise reduction in multimodal image fusion task to present a showcase. Most modal-specific denoise methods focus on aggregating useful information from multimodal data and mitigating the influence of noise in multimodal fusion.

\subsubsection{Joint variation based fusion}
Joint optimization is often used in variation-based noise reduction when fusing multiple visual modalities, e.g., RGB and thermal. In unimodal noise reduction field, taking image modality as an example, classical denoise model based on Total Variation (TV) is equivalent to solving the following optimization problems:
\begin{equation}
   \min  \iint\left [ u(x, y) - u_0(x, y) \right ] ^2dxdy + \lambda \iint \left | \bigtriangledown u(x, y)  \right |  dxdy,  
\end{equation}
where $u_0(x, y), u(x, y)$ represent the input pixel and output pixel after noise reduction respectively. For multimodal noise reduction, the joint variational model for fusion and denoise is defined as follows:
\begin{equation}
    \min  \iint \sum_{m = 1}^{M} w_m \left [ u(\boldsymbol{x} ) - u_m(\boldsymbol{x} ) \right ] ^2d\boldsymbol{x} + \lambda \iint \left | \bigtriangledown u(\boldsymbol{x} )  \right |  d\boldsymbol{x},  
\end{equation}
where $u_m(\boldsymbol{x} )$ represents the input data from modality $m$, $w_m$ is the non-negative weight satisfying $\sum_{m = 1}^{M}w_m=1$. In a nutshell, joint optimization is the solution of a specific optimization problem using data from different modalities. Wang \emph{et al}. \cite{wang2008variational} proposed a variational model in the pixel domain and wavelet domain for joint fusion and denoise of noisy multifocus images. Kumar \emph{et al}. \cite{kumar2009total} also employed the total variational model to fuse images acquired using multiple sensors and achieves excellent results on two modal images provided in the medical field and aircraft navigation field. Recently, Padmavathi \emph{et al}. \cite{padmavathi2020novel} proposed an image fusion algorithm for constructing a fused image using the total variation model. They incorporated an optimized adaptive weighting fusion scheme for medical multimodal data analysis (i.e., Magnetic Resonance images and Positron Emission Tomography images). Nie \emph{et al}. \cite{nie2021total} introduced a total variation-based fusion method for infrared and visible images. The method leverages the total variation model to fuse two modalities, enabling enhanced image quality and information integration. Quan \emph{et al}. \cite{quan2021relative} developed a feature fusion method called relative total variation structure analysis (RTVSA) for urban area classification. This approach combines various features derived from hyperspectral imaging (HSI) and LiDAR data, providing a comprehensive analysis for accurate classification. Furthermore, Liu \emph{et al}. \cite{liu2022tse_fuse} designed a two-stage enhancement (TSE) framework for infrared and visible image fusion. Their framework incorporated an attention mechanism and a feature-linking model (FLM), combining structure adaptive total-variational (SATV) and L1 sparsity terms to extract two-scale detail layers and a base layer. The proposed method demonstrates robustness and effectiveness in improving the quality of fused images.

\subsection{Cross-modal noise reduction}
Many multimodal tasks (e.g., multimodal object detection, vision-and-language understanding) highly depends on correctly aligned multimodal training data. However, the real-world multimodal pairs tend to contain weakly aligned or even unaligned samples~\cite{changpinyo2021conceptual}. For example, in RGB-thermal multimodal object detection, multimodal inputs are often weakly aligned, i.e., the locations of the same objects may be shifted across different modalities~\cite{zhang2019weakly}. In social media, the text descriptions are frequently irrelevant with the image content, i.e., unpaired modalities. In this section, we consider weakly aligned or unaligned multimodal samples as cross-modal noise. Compared to modal-specific noise, cross-modal noise lies in a higher-level semantic space. Current cross-modal noise reduction methods can be roughly split into rules-based filtering, model-based rectifying and noise robustness regularization.

From the data perspective, some strict rules are used to conduct data cleaning~\cite{sharma-etal-2018-conceptual, radenovic2023filtering, gadre2023datacomp}. As a recent representative method, CAT (Complexity, Action, and Text-spotting) is a filtering strategy which is devised to select informative image-text pairs, thereby reducing the influence of cross-modal noise~\cite{radenovic2023filtering}.  In multi-spectral  object detection, image registration (i.e., spatial alignment) is a commonly used pre-processing~\cite{maintz1998survey,zitova2003image}. By employing geometric rules to align two images, the shifted positions can be rectified across different modalities.

From the model perspective, model filtering or rectifying methods try to identify cross-modal noise samples and further remove or rectify them. To achieve this, Huang \emph{et al}. proposed a method called Noisy Correspondence Rectifier (NCR) to tackle cross-modal noise~\cite{huang2021learning}. NCR leverages the memorization effect of neural networks, dividing the data into clean and noisy subsets based on loss differences. Subsequently, it rectifies correspondence using an adaptive prediction model in a co-teaching manner. Applied to image-text matching as a showcase, NCR achieves superior performance in the presence of cross-modal noisy data. ALBEF~\cite{li2021align} adopts momentum models to generate pseudo targets as additional supervision. BLIP~\cite{li2022blip} introduces a filter to remove the noisy data according to the similarity of image-text pairs and then utilizes a captioner to regenerate the corresponding web texts. This filtering and captioning strategy can improve the quality of multimodal pairs, contributing to the improvement of downstream vision-language tasks.

Noise robust regularization is another perspective to mitigate the influence of cross-modal noise. To explicitly stabilize and harmonize multimodal pretraining and mitigate the influence of latent noise, NLIP~\cite{huang2023nlip} employs noise-adaptive regularization to avoid overfitting to noisy image-text pairs, adjusting alignment labels based on estimated noisy probabilities. Additionally, NLIP utilizes a concept-conditioned cross-modal decoder to generate synthetic captions to impute the missing object information. To mitigate the influence of cross-modal noise in vision-language task, Li \emph{et al}.~\cite{li2020oscar} proposed OSCAR that detects object tags in images and uses them as anchor points for alignment. Recent work provides both theoretical and empirical analysis to shed light on the influence of cross-modal noise in multimodal contrastive learning~\cite{nakada2023understanding}. Furthermore, the research introduces a novel MMCL loss designed to handle unpaired multimodal samples, demonstrating enhanced robustness in empirical experiments.

\subsection{Discussion}
To summarize, learning from noisy multi-modal data is a common but challenging problem. Current approaches address this problem from two perspectives: modal-specific noise reduction (for feature noise) and cross-modal noise reduction (for semantic noise). However, these methods typically focus on specific scenarios, such as multimodal image fusion or automatic driving, leaving the exploration of general noise patterns and learning paradigms relatively unexplored. We have identified several potential research topics in this field. First, it is important to leverage the correlation between noises in different modalities. For instance, images with similar wavelengths in a hyper-spectral image often exhibit similar noise patterns. Second, it will be effective to leverage the complementarity between noisy and clean modalities for noise reduction. Third, addressing high-level semantic noise poses an interesting direction and it is more challenging. For example, how can we tackle this problem using Multimodal Large Language Models (MLLM)?
\input{table/noise}

%% file: table/noise.tex
\begin{table*}[ht]
\centering
\caption{A summary of representative multi-modal noise reduction works.}
\begin{adjustbox}{max width=1\textwidth}
\begin{tabular}{@{}|c|c|c|c|c|@{}}
\toprule
\textbf{Noise type} & \textbf{Reference} & \textbf{Denoise Paradigm} & \textbf{Application}                           \\ 
        \midrule

    \multirow{6}{*}{Modal-specific} &

        \cite{kumar2009total}     & \multirow{6}{*}{Joint variation} &\multirow{6}{*}{Image fusion}   \\ \cline{2-2}   &
        
        \cite{zhao2017medical}   &   &  \\ \cline{2-2}   &
        
        \cite{padmavathi2020novel} &  &   \\ \cline{2-2}   &
        
        \cite{nie2021total}   &  & \\ &
        
        \cite{liu2022tse_fuse}    &  &    \\ \cline{2-2}  \cline{4-4} &
        
        \cite{quan2021relative}    &  & Remote-sensing scene classification  \\ \midrule \multirow{5}{*}{Cross-modal} & \cite{sharma-etal-2018-conceptual, radenovic2023filtering, gadre2023datacomp}
        
         &\multirow{1}{*}{Rules-based filtering}  & \multirow{1}{*}{Multimodal pretraining}   \\\cline{2-4}  &\cite{huang2021learning} 
        
         &\multirow{2}{*}{Model-based rectifying}  &Image-text matching   \\ \cline{2-2} \cline{4-4}  &\cite{li2021align,li2022blip} 
        
         &  &Multimodal pretraining \\\cline{2-4}  &\cite{huang2023nlip,nakada2023understanding} 
        
         &\multirow{2}{*}{Noise robust regularization}  &Multimodal pretraining   \\\cline{2-2} \cline{4-4}  &\cite{li2020oscar} 
        
         &  &Vision-language understanding   \\

    \bottomrule
\end{tabular}
\end{adjustbox}
\end{table*}

%% file: incomplete/incomplete.tex
\section{Incomplete multimodal Learning}
\begin{figure*}[!t]
    \centering
    \includegraphics[width=0.99\textwidth]{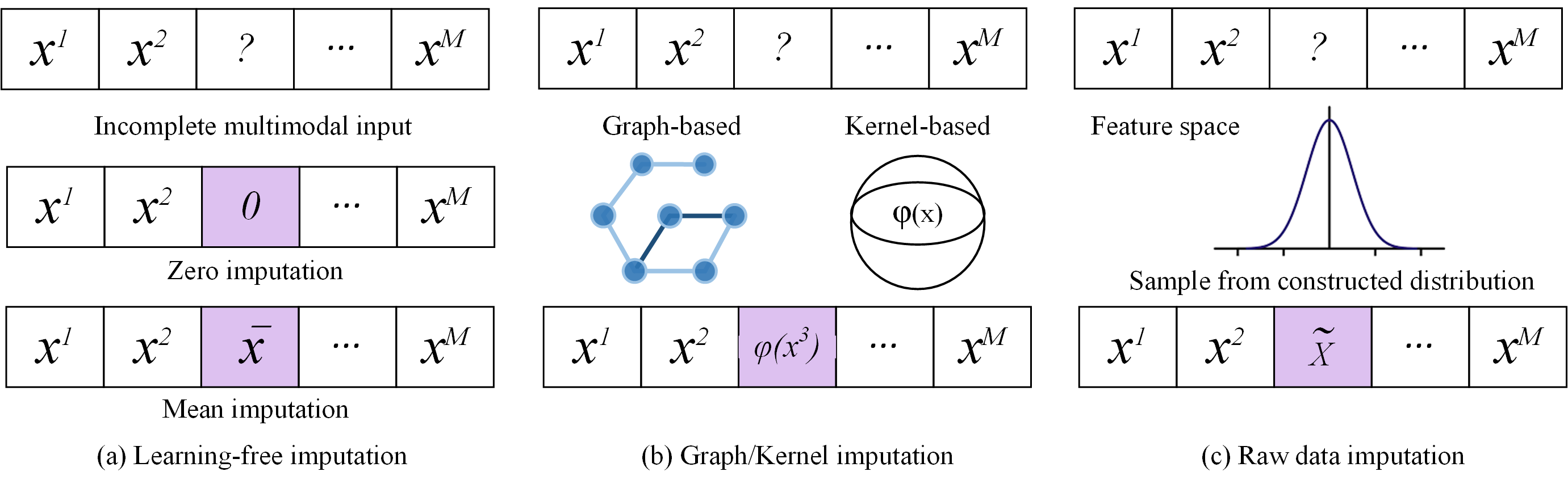}
    \caption{\textbf{Imputation based incomplete multimodal learning.}} 
    \label{fig:incomdata}
\end{figure*}

The collected multimodal data in real applications are often incomplete with partial missing modalities for some samples due to unexpected factors, such as equipment damage, data transmission and storage loss. For a concrete example, in recommendation system, the browsing behavior history and credit score information may not always be available for some users \cite{shao2013clustering}. Similarly, multimodal data can be highly incomplete in medical analysis due to the patients' personal preferences, economic constraints and limication in medical resources. For example, although combining data from multiple modalities e.g., magnetic resonance imaging (MRI) scans, positron emission tomography (PET) scans, and cerebrospinal fluid (CSF) information can obtain a more accurate diagnosis for Alzheimer's Disease \cite{zhang2012multi,liu2014inter}, owing to high measurement cost for PET scans and uncomfortable invasive tests for CSF, some patients may refuse to conduct these examinations. Thus, incomplete multimodal data commonly exist in Alzheimer's Disease diagnosis \cite{thung2014neurodegenerative}. Typically, conventional multimodal learning models assume the completeness of multimodal data and thus cannot be applied to partial missing modalities directly. For this issue, incomplete multimodal learning, aiming at exploring information of incomplete multimodal data with partial missing modalities, appears and has received increasing research interests in recent years \cite{wen2022survey}. In this section, we mainly focus on the current progress of researches on incomplete multimodal learning. From the view of whether to impute the missing data or not, we categorize existing methods into two groups including imputation-based and imputation-free incomplete multimodal learning, where the imputation based methods are further divided into two groups as shown in Fig.~\ref{fig:incomdata} including instance and modality level imputation.

\subsection{Imputation based incomplete multimodal learning}
For the incomplete multimodal data, a natural approach to address the missing modality learning issue is to directly impute the missing modalities or related missing information in the kernels or graphs constructed from the data as shown in Fig.~\ref{fig:incomdata}. Then the conventional multimodal learning algorithm can be applied to downstream tasks. In this branch, the existing imputation-based works can be further divided into two categories, including model-agnostic imputation and model-specific imputation. The model-specific imputation methods generally design a imputation model or network for missing data recovery while the imputation model is not necessary for the model-agnostic imputation methods which usually utilizes heuristic filling strategy for the missing modalities.

\subsubsection{Model-agnostic imputation}
For the missing modalities, the simplest imputation schemes are zero imputation and mean value imputation, i.e., filling missing values with zero or imputation with the averaged value in the corresponding modality, which is widely used as the baseline methods in most studies \cite{shao2015multiple,zhao2016incomplete}. However, previous work~\cite{shao2015multiple} shows that the naive imputation may lead to unpromising performance. To address this issue, a weighted matrix factorization model is proposed to assign small weights for the imputed modalities, mitigating the negative influence on optimization \cite{shao2015multiple}. Furthermore, for an online multimodal clustering case, a dynamic weighting and imputation mechanism is designed according to the valid modalities and task at hand\cite{shao2016online}. To implement the graph-based approach on incomplete multimodal data, Wang \emph{et al}.~\cite{wang2019spectral} utilize similarity values computed from valid modalities to impute the missing elements of graph. In \cite{gao2016incomplete}, the missing elements in the incomplete kernels are pre-imputed by the average of all columns. In \cite{zhou2019consensus}, the authors also utilize the average instance of the corresponding modality to impute the missing instances and then assign a small weight to those imputed instances during the graph construction phase.

\subsubsection{Learning based imputation}
For the learning based imputation, according to the filling values, we can further divide it into two categories: kernel/graph-based imputation and raw data imputation.

\textbf{Kernel/graph-based imputation}: Kernel and graph learning based approaches are two main lines of incomplete multimodal learning, which transform the learning phase from the raw data to the kernel space or graph. For the kernel and graph constructed from each modality, each element represents the relationship in kernel space or affinity/similarity of instance pairs. The partial missing instances in each modality result in incomplete kernels and graphs. A line of works has been proposed to address the incomplete kernel/graph-based multimodal learning issue. Here we highlight a few seminal works. David Williams \emph{et al}.~\cite{williams2005analytical} proposed a Gaussian mixture model based method that can analytically compute the missing Gaussian kernel elements associated with the missing instances. Trivedi \emph{et al}.~\cite{trivedi2010multiview} proposed a kernelized canonical correlation analysis (KCCA) based imputation method, which minimizes a Laplacian kernel regularization to obtain the missing kernel elements. Although effective, it is only applicable to the two-modal data with one complete modality. Lately, collective kernel learning (CKL)~\cite{shao2013clustering} is proposed to mutually complete the two incomplete kernels associated with the two incomplete modalities. For the data with more than two modalities, such as the case with $k$ modalities ($k \ge 3$), CKL needs to recover each kernel matrix ${K_{i + 1}}$ of the $\left( {i + 1} \right)$th modality according to kernel ${K_i}$ in a cycle iteration scheme. Sahely \emph{et al}.~\cite{bhadra2017multi} proposed a multimodal kernel completion method based on the sparse reconstruction technique. The method can complete the missing rows and columns of all kernels jointly in a within-kernel preservation and between-kernel exploration framework. Yang \emph{et al}.~\cite{yang2018semi} reduced incomplete multi-modal imputation into semi-supervised learning problem. They further proposed SLIM (Semi-supervised Learning with Incomplete Modality ) and its kernel vision SLIM-K based on matrix completion for solving the modal incompleteness. Recently, Liu \emph{et al}. proposed several flexible kernel methods to recover the missing kernel elements for incomplete multimodal clustering tasks~\cite{liu2019multiple,ye2017consensus,zhu2018localized,liu2021incomplete}. 

By aligning all incomplete kernels to a consensus kernel derived from the consensus orthogonal representation, these methods can jointly obtain the missing elements associated with missing instances and the consensus partition representation shared by all modalities for incomplete multimodal data clustering. Besides of kernel-based methods, an adaptive graph completion method is developed to recover the missing similarities of all graphs with respect to all modalities~\cite{wen2020adaptive}. The method mainly employs the sparse-representation-based data reconstruction technique to borrow the graph information of other modalities in recovering the missing elements of each modality.

\textbf{Raw data imputation}: Some methods recover the missing modalities at raw feature level. For example, VIGAN (View Imputation via Generative Adversarial Network)~\cite{shang2017vigan} is a pioneering work based on generative adversarial network (GAN) and Autoencoder for missing modality recovery. VIGAN employs GAN to initialize the missing modality according to another modality of the same sample and then uses denoise Autoencoder to refine the recovered missing modalities. A limitation of VIGAN is that it is not applicable to data with more than two modalities. Subsequently, partial multimodal clustering via consistent GAN (PMVC\_CGAN) \cite{wang2018partial} and adversarial incomplete multimodal clustering (AIMC) \cite{xu2019adversarial} are proposed for missing modality imputation and incomplete multimodal clustering. Different from VIGAN which uses one available data to generate the corresponding missing modality, PMVC\_CGAN and AIMC seek to learn the missing modalities from the common representation encoded by the un-missing modalities of the same sample. In~\cite{arya2021generative}, a two-stage incomplete deep multimodal network is proposed for breast cancer prediction. Similar to the network structures of AIMC and PMVC\_CGAN, GAN is introduced to enhance the quality of missing view imputation. Besides of GAN-based methods, CRA (cascaded residual autoencoder) stacks all modalities into one modal and treats the missing modalities imputation as conventional missing feature completion by optimizing a single-modal cascaded residual autoencoder network \cite{tran2017missing}. Liu \emph{et al}.~\cite{Liu2023Information} proposed another representative autoencoer based imputation framework termed RecFormer for clustering task. Specifically, RecFormer employs a two-stage autoencoder network with the self-attention structure to synchronously extract high-level semantic representations and recover the missing data

To obtain a more reliable modality imputation performance, Tang \emph{et al}.~\cite{tang2022deep} proposed to impute the modality as the average of corresponding semantic neighbors. Lin \emph{et al}.~\cite{lin2022dual} established a dual prediction model that can predict the latent representation of the missing modality according to that of the observed modality and then used the decoder to obtain the recovered missing modality according to the predicted latent representation. matrix factorization is also applied for raw modality imputation. For example, UEAF (Unified Embedding Alignment Framework)~\cite{wen2019unified} and IMCRV (Incomplete multimodal Clustering with Reconstructed Views)~\cite{yin2021incomplete} provide two matrix factorization based missing modality recovery models, which can reversely reconstruct the data of missing modalities from the common representation. In~\cite{wen2021unified}, low-rank representation and tensor learning are introduced for joint missing modality recovery and graph completion. Explicitly taking the reliability of imputation into account, Xie \emph{et al}.~\cite{xie2023exploring} introduced the Uncertainty-induced Incomplete Multi-View Data Classification (UIMC) model to address challenges in classifying incomplete multi-view data with missing perspectives. UIMC explores and exploits uncertainty by imputing missing modalities as distributions rather than deterministic values, allowing for more reliable predictions. The model employs an evidential multimodal learning framework to weigh imputed data, reducing the impact of low-quality imputations. In incomplete multimodal action recognition task, Sangmin \textit{et al.} proposed a modular network called ActionMAE, which randomly drops modality features and then learn to impute them.

Lin \emph{et al}. proposed a novel objective by integrating representation learning and data recovery into a unified information-theoretic framework termed COMPLETER~\cite{lin2021completer} for clustering task. Specifically, COMPLETER propose to learn intra-modal consistency by maximizing mutual information and learn modal imputation by minimizing conditional entropy. Dong \emph{et al}. proposed a deep Gaussian processes model to impute the missing modalities and further enhance the performance of downstream tasks~\cite{dong2023partial}. Motivated by contrastive learning, Yang \emph{et al}. introduced a multimodal imputation paradigm termed SURE for robust multimodal imputation~\cite{yang2022robust}. The method treats complete multimodal pairs as positive and randomly sampling unpaired samples as negative. A noise-robust contrastive loss is employed to alleviate the influence of false negatives. Li \emph{et al}. proposed a prototype-based multimodal imputation method, which incorporates a two-stream model with dual attention layers and contrastive learning mechanism to learn modal-specific prototypes and model the cross-modal relationship~\cite{li2023incomplete}. Considering incomplete multimodal inputs, the model performs modality imputation using prototypes learned from the incomplete multimodal training data and the sample-prototype relationship inherited from the observed modalities. Xu \emph{et al}.~\cite{ren2024novel} proposed a novel federated multi-view clustering method using an unsupervised technique to evaluate and refine imputation quality, efficiently handling various scenarios of incomplete multi-view data.

\subsection{Imputation-free incomplete multimodal learning}
Different from the above imputation based methods, imputation-free methods only focus on exploring the information corresponding to the available modalities. Most of the related imputation-free methods can be grouped into four mainstreams: latent representation and projection learning, graph learning, kernel learning, and deep learning.
\subsubsection{Latent representation and projection learning}
Latent representation and projection learning based incomplete multimodal learning (taking clustering for example) generally seek to obtain a multimodal latent representation by exploring the available partially aligned information among these available modalities. In this branch, partial multimodal clustering (PMVC)~\cite{li2014partial} is a popular and pioneering work based on matrix factorization. It concentrates on encoding the data with two fully observed modalities into a common representation within the latent subspace. Based on PMVC, incomplete multimodal grouping (IMG) \cite{zhao2016incomplete} and partial multimodal subspace clustering \cite{xu2018partial} further introduce the graph constraints to capture the structure information in the latent common subspace. A basic model used in the above methods to learn the latent common representation from the incomplete bi-modal data.

Zhou \emph{et al}.~\cite{zhou2019latent} integrated the latent representation learning and regression into a unified framework for Alzheimer's disease diagnosis based on incomplete multimodal neuroimaging and genetic data. However, a widely recognized limitation of the above methods based on model (4) is that they are only applicable to the case with bi-modal data. To address this limitation, weighted matrix factorization is introduced, where doubly aligned incomplete multimodal clustering \cite{hu2018doubly}, one-pass incomplete multimodal clustering \cite{hu2019one}, and localized sparse incomplete multimodal clustering \cite{liu2023localized} are representative methods. These models generally impose the location information of missing instances as the weighted matrices with binary values on the matrix factorization model to eliminate the loss yet reduce the negative influence of the missing instances. The basic model of the weighted matrix factorization based methods can be formulated as follows:
\begin{equation}\label{im_eq3}
\mathop {\min }\limits_{{U^{(m)}},P} \sum\limits_{m = 1}^M {\left\| {{W^{(m)}}\left( {{X^{(m)}} - P{U^{(m)}}} \right)} \right\|_F^2 + \Psi \left( {{U^{(m)}},P} \right)},
\end{equation}
where ${{X^{(m)}}}$ denotes the data of the $m$-th modality with missing instances being 0 or the average value of the available instances in the modality. ${{U^{(m)}}}$ denotes the basic matrix and ${\Psi \left( {{U^{(m)}},P} \right)}$ denotes the constraints of these variables. $P$ is the latent complete common representation to learn. $W^{(m)}$ is a diagonal matrix constructed according to the instance missing location information in the $m$-th modality. For example, if the $i$-th sample does not have the $m$-th modality, then $W_{i,i}^{(m)}$ can be simply set as 0, otherwise $W_{i,i}^{(m)}=1$. With such a setting, the negative influence of missing modalities can be eliminated from the learning model, and the available aligned consistent and complementary information can be explored for the latent common representation learning.

Besides of these weighted matrix factorization models, a generalized incomplete multimodal clustering model is proposed to learn a complete common representation from the incomplete modal-specific representations with a partially aligned learning term \cite{wen2020generalized}. Specifically, such a partially aligned learning term can be regarded as a variant of the weighted common representation learning. To prevent the model from overly focusing on certain modalities with larger dimensions, projective incomplete multimodal clustering transforms the matrix factorization into a variant of the weighted projection learning model to obtain the consensus representation from the incomplete multimodal data \cite{deng2023projective}. For the incomplete multimodal and incomplete multi-label classification task, a variant of the weighted projection learning model \cite{li2021concise} and weighted matrix factorization model \cite{tan2018incomplete} are developed to explore the information of available modalities. In terms of infant neuroscience, Zhang \emph{et al}. propose to explore incomplete multimodal data by constructing a latent representation to capture the complementary information from incomplete time-points~\cite{zhang2018infant}. In summary, the above methods focus on exploring the available information of the observed modalities by introducing some techniques like the weighting mechanism to eliminate the loss of those missing modalities.

\subsubsection{Graph learning based}
Graph learning based method is very popular in conventional multimodal clustering. This technique is also widely considered and integrated into other methods like matrix factorization based and deep learning based methods, to explore the structure information of data so as to enhance the performance. For incomplete multimodal learning, the main challenges to apply the graph learning based approach are two-folds: (1) How to obtain the intrinsic structure relationships among the available instances in each modality? (2) How to fully explore these incomplete relationships? For example, for the incomplete multimodal clustering tasks, one of the major challenges is to obtain the consensus complete graph or representation from these incomplete modal-specific relationships associated with all incomplete modalities. Specifically, we can formulate this issue as the following simple framework:
\begin{equation}\label{im_eq4}
\mathop {\min }\limits_{\left\{ {{S^{\left( m \right)}}} \right\}_{m = 1}^l,Z} \sum\limits_{m = 1}^l {f\left( {{{\bar X}^{\left( m \right)}},{S^{\left( m \right)}}} \right) + g\left( {{S^{\left( m \right)}},{W^{\left( m \right)}},Z} \right)},
\end{equation}
where ${{{\bar X}^{\left( m \right)}}}$ denotes the set of the available instances in the $m$-th modality. ${{S^{\left( m \right)}}}$ represents the modal-specific graph learned from the $m$-th modal data ${{{\bar X}^{\left( m \right)}}}$. ${{W^{\left( m \right)}}}$ is the modal-available or missing indicator matrix as defined in Eq.~\ref{im_eq3}. Taking the incomplete multimodal clustering task for example, the target of general graph learning based method is to learn the consensus graph or representation $Z$ from the incomplete data $\left\{ {{{\bar X}^{\left( m \right)}}} \right\}_{m = 1}^l$ or graph $\left\{ {{S^{\left( m \right)}}} \right\}_{m = 1}^l$, such that a consensus clustering result can be obtained by performing k-means or spectral clustering on $Z$. 

As one representative work, incomplete multimodal spectral clustering~\cite{wen2018incomplete} proposed a unified framework that integrates the adaptive collaborative representation based modal-specific graph learning and co-regularized consensus representation learning. The first term enables the method to adaptively capture the intrinsic relationships among available instances of each modality and the second term seeks to obtain the consensus complete representation by fully exploring such similarity relationship. For the data with only two modalities, Wu \emph{et al}.~\cite{wu2018incomplete} proposed a method based on structured graph learning, which seeks to learn two modal-specific graphs corresponding to the two modalities with a partially aligned constraint on the samples with complete modalities. Then, the method obtains the modal-specific clustering from the two graphs and adopts the best one-to-one map to obtain the final consensus clustering result. The work in \cite{wen2023highly} pre-constructs several modal-specific graphs from the available instances of every modality and then learns a consensus graph from these graphs via a weighted consensus graph learning model with the constraint of pre-constructed confident neighbor information. In summary, except for the imputation based graph methods, most graph-based incomplete multimodal learning methods seek to exploit certain similarity information among available instances to discover the clustering structure and clustering result. In \cite{xiang2013multi}, for Alzheimer's disease prediction, a unified linear regression based model is developed, in which the incomplete multimodal data are segmented into several complete subsets for model optimization. Similar to the above method, Liu \emph{et al}.~\cite{liu2017view} segmented the incomplete multimodal data into several complete subsets and used the sparse representation algorithm to construct the sub-graphs from these subsets. Then a modal-aligned hypergraph learning model is developed to perform the Alzheimer's disease diagnosis.

\subsubsection{Kernel learning based}
Kernel learning based methods transform the learning phase from the raw-feature space into the kernel space. Similar to the graph learning based methods, each element of the kernel matrix denotes a kind of the similarity or connection values of the corresponding two samples. For incomplete multimodal data, the constructed kernels w.r.t. different modalities will also be incomplete with missing rows and columns. Thus, the target of kernel learning based methods for incomplete multimodal data is to explore the information of these incomplete kernels. Late fusion incomplete multimodal clustering~\cite{liu2018late} is one of the representative kernel imputation-free methods for incomplete multimodal data. The method does not need to impute the missing kernel elements but tries to simultaneously impute the missing representations corresponding to the missing modalities and learn the consensus representation in a joint framework. Furthermore, to reduce the low-quality imputation of the latent representations, Liu \emph{et al}.~\cite{liu2020efficient} proposed an efficient and effective regularized incomplete multimodal clustering method (EERIMVC), which integrates the initialized consensus representation as prior knowledge. In this work, such prior representation is obtained by the conventional complete kernel algorithm with zero or average filling. Therefore, the quality of the initialized consensus representation will be an important factor to the performance. To break the gap between representation learning and label prediction, a one-stage incomplete multimodal clustering method is proposed \cite{zhang2021one}. Compared with EERIMVC, this method replaces the consensus representation with a clustering label matrix and an orthogonal matrix.

For the above late fusion based kernel methods, we can observe that: (1) These methods can simultaneously recover the modal-specific missing representations and learn the consensus representation by exploring the partially aligned structure information of data. (2) The performance of these methods highly relies on the quality of the initialized incomplete basis representations derived from the incomplete kernels.

\subsubsection{Deep learning based}
Recent years, deep learning received lots of attention in incomplete multimodal learning. For instance, Zhao \emph{et al}.~\cite{zhao2018incomplete} proposed a deep partial multimodal clustering network, which first uses deep encoders to extract the latent features of multimodal data and then adopts the graph regularized partial multimodal clustering model to learn the consensus representation from the incomplete bi-modal data. For data with more than two modalities, Wen \emph{et al}.~\cite{wen2020dimc} proposed several deep incomplete multimodal clustering networks based on deep autoencoder framework, cognitive learning \cite{2020CDIMC}, and graph neural network \cite{wen2021structural}, respectively. The objective function of these methods can be unified as the following formula:
\begin{equation}\label{im_eq6}
\min \sum\limits_{m = 1}^l {f\left( {{X^{\left( m \right)}},{{\bar X}^{\left( m \right)}}} \right)}  + {\lambda}g\left( {Z,\left\{ {{Z^{\left( m \right)}}} \right\}_{m = 1}^l} \right),
\end{equation}
where ${X^{\left( m \right)}} \in {R^{n \times {d_m}}}$ and ${\bar X^{\left( m \right)}} \in {R^{n \times {d_m}}}$ denote the original and reconstructed $m$-th modal data. $n$ denotes the number of samples and ${d_m}$ is the feature number of the $m$-th modal sample. ${W^{\left( m \right)}} \in {\left\{ {0,1} \right\}^{n \times n}}$ is a diagonal binary matrix as defined in (\ref{im_eq3}) to indicate the location of the missing modalities. $\bar Z$ and $\left\{ {{Z^{\left( m \right)}}} \right\}_{m = 1}^l$ represent the consensus or fused latent representation and the modal-specific latent representation at one layer of the network. ${\lambda}$ is penalty parameter. $f$ and $g$ are functions or constraints for these variables. For example,  $f$ is set as the mean square error (MSE) loss function in many methods. Objective (\ref{im_eq6}) can be also split into two independent losses to optimize two networks separately, where the first loss with the function $f$ is generally used for network pre-training. Similar to these unsupervised multimodal learning methods, Liu \emph{et al}. \cite{liu2023dicnet} developed an incomplete multimodal multi-label classification network, with the autoencoder based framework and multimodal contrastive constraint.

The above methods focus on the observed instances to train the network and can flexibly handle all kinds of incomplete data with more than two modalities by introducing the location matrix ${W^{\left( m \right)}}$ as prior weight in function $f$. In addition, a common technique adopted in these three methods is the weighted latent representation fusion, in which the location information of missing modalities is adopted as fusion weight to reduce the negative influence of missing modalities. Lately, Xu \emph{et al}.~\cite{xu2022deep} proposed an imputation-free and fusion-free deep network for incomplete multimodal data clustering. The method mines the multimodal cluster complementarity in the subspace and transforms such complementary information as supervised information to guide the network training. However, this method needs a part of samples with fully observed modalities for model training.

For supervised learning tasks, a incomplete multimodal learning framework is proposed in \cite{zhang2019cpm}, which can flexibly learn a common representation for arbitrary missing data and preserve both common and complementary information. For the more difficult task with incomplete labels, a deep double incomplete multimodal multi-label learning network~\cite{wen2023deep} is proposed. This deep network mainly introduces two mask constraints \textit{w.r.t.} the missing view locations and missing label locations, to reduce the negative impact of the missing labels and missing views. A good property of the method is that it can handle all kinds of incomplete data classification tasks including supervised/semi-supervised cases, complete/incomplete data, and complete/incomplete label cases. More recent work~\cite{mckinzie2023robustness} focuses on an even more generalized scenario than incomplete multimodal learning which is so called \textit{train-test modality mismatch}, where the test multimodal data can contains more or less modalities than training data. Specifically, all modalities are available at pretraining stage, while a completely different subset is encountered by the model during supervised training and testing in downstream tasks.

In medical image processing, Sanaz \textit{et al.} proposed MMCFormer~\cite{karimijafarbigloo2024mmcformer}, a novel missing modality compensation network to address the incomplete modalities for brain tumor segmentation. The proposed method is build upon 3D efficient transformer blocks and uses a co-training strategy to effectively train a missing modality network by ensuring feature consistency between complete modalities and incomplete modalities. Yao \textit{et al.} proposed DrFuse~\cite{yao2024drfuse} for effective medical diagnosis and forecasting prognosis with incomplete multimodal data. DrFuse first learns the common features shared between all modalities and modality-specific features within each separated modality and then weight and then combining the prediction from each modality to make the final prediction. Specifically, a disease-wise attention layer is devised to produce the patient- and disease-wise weights to tackle the modality inconsistency issue.

Some representative incomplete multimodal learning methods are summarized in Table 4.

\subsection{Discussion}
To sum up, the ways to handle incomplete data can be commonly divided into two categories, imputation-based incomplete multimodal learning and imputation-free incomplete multimodal learning. Methods that focus on imputing incomplete modalities can be classified into two types: one indirectly recovers the missing modalities through graph or kernel completion, and the other directly fills in the original data.
For imputation-free incomplete multimodal learning methods, their design philosophy focuses more on utilizing available modalities to explore information and understand multimodal data. However, both types of methods for handling incomplete multimodal data face deeper challenges. For example, the quality assessment of imputed instances for missing modalities is commonly overlooked. Besides, using prior missing information to mask unknown modalities is itself difficult to bridge the information gap and information imbalance issues caused by modal missing. Moreover, incomplete cases such as unpaired views \cite{wang2022lung,yu2020flexible,wang2024partially} and partially missing labels \cite{xing2021weakly} also need more attention. 
\input{incomplete/incomplete-table}

%% file: incomplete/incomplete-table.tex
\begin{table*}[htbp]
\setlength{\abovecaptionskip}{0.0cm}  
\setlength{\belowcaptionskip}{0.0cm} 
\vspace{-0.0cm}
\begin{center}
\caption{A summary of some representative incomplete multimodal learning works.}
\center\resizebox{0.99\textwidth}{!}{
{
\setlength{\tabcolsep}{1.9mm}
\begin{tabular}{c|c|c|c}
\toprule
   \multirow{1}{*}{\textbf{Category}}      & \multirow{1}{*}{\textbf{Base model}} &
\multirow{1}{*}{\textbf{Task}} &\multirow{1}{*}{\textbf{Reference}}
\\
\midrule

\multirow{2}{*}{Learning-free imputation}     &Matrix factorization & \multirow{6}{*}{Clustering} &   \cite{shao2015multiple},\cite{shao2016online}
            \\ \cline{2-2} \cline{4-4} 
        &\multirow{2}{*}{Graph learning}&  &\cite{wang2019spectral},\cite{gao2016incomplete}, \cite{zhou2019consensus}
            \\ \cline{1-1} \cline{4-4}
        & & &\cite{wen2020adaptive} \\ \cline{2-2} \cline{4-4}
   \multirow{12}{*}{Learning based imputation}   &\multirow{3}{*}{Kernel learning} &  & \cite{trivedi2010multiview} 
 \\ \cline{4-4}
     & & & \cite{shao2013clustering,yang2018semi} 
            \\ \cline{4-4}
       & &  & \cite{bhadra2017multi} \cite{liu2019multiple}, \cite{zhu2018localized}, \cite{liu2021incomplete}
              
            \\ \cline{2-4}
          &\multirow{7}{*}{Raw data imputation} & \multirow{4}{*}{Modality generation} & \cite{shang2017vigan}
  \\ \cline{4-4}
    & &  &  \cite{wang2018partial} 
    \\ \cline{4-4}
     & &  & \cite{xu2019adversarial} 
\\ \cline{4-4}
  & & &  \cite{tran2017missing,Liu2023Information,tang2022deep} 
    \\ \cline{3-3} \cline{4-4}
     & & Breast cancer prediction &  \cite{arya2021generative}
  \\ \cline{3-3} \cline{4-4}
    & & Representation learning &  \cite{dong2023partial} 
   \\ \cline{3-3} \cline{4-4} & & Action recognition &\cite{woo2023towards}
  \\ \cline{2-4}
  &\multirow{4}{*}{Matrix factorization}& \multirow{5}{*}{Clustering} & \cite{wen2019unified,yin2021incomplete}
  \\ \cline{4-4}
   & &  &  \cite{wen2021unified}
  \\ \cline{1-1} \cline{4-4}
\multirow{16}{*}{Imputation-free}   & &  &\cite{li2014partial},\cite{zhao2016incomplete},\cite{xu2018partial} 
  \\ \cline{4-4}
  & &  &  \cite{hu2018doubly},\cite{hu2019one} 
  \\ \cline{2-4}
   &\multirow{3}{*}{Projection learning} &  \multirow{2}{*}{Regression} &\cite{deng2023projective} 
  \\ 
 &    & &  \cite{zhang2018infant}
  \\ \cline{3-3} \cline{4-4}
   & & Classification & \cite{li2021concise}
  \\ \cline{2-4}
 &Linear regression  & \multirow{2}{*}{Alzheimer's disease prediction} &  \cite{xiang2013multi}
  \\ \cline{2-2} \cline{4-4} 
  &Representation learning &  & \cite{zhou2019latent}
  \\ \cline{2-4}
     &\multirow{3}{*}{Graph learning} & \multirow{2}{*}{Clustering} & \cite{wu2018incomplete} 
  \\ \cline{4-4}
  & &  & \cite{wen2018incomplete}, \cite{wen2023highly}  
 \\ \cline{3-3} \cline{4-4}
    & & Alzheimer's disease diagnosis &  \cite{liu2017view}
  \\ \cline{2-4}
   &Kernel learning & \multirow{3}{*}{Clustering} & \cite{liu2018late}, \cite{liu2020efficient}, \cite{zhang2021one} 
  \\ \cline{2-2} \cline{4-4}
    &Matrix factorization &  & \cite{zhao2018incomplete}
  \\ \cline{2-2} \cline{4-4}
   &Weighted fusion &  & \cite{2020CDIMC}, \cite{wen2021structural}, \cite{xu2022deep}  
  \\ \cline{2-4} 
   &Representation learning strategy& \multirow{2}{*}{Classification} &  \cite{zhang2019cpm,mckinzie2023robustness,yao2024drfuse}\\ \cline{2-2} \cline{4-4}
   &Weighted fusion&  & \cite{wen2023deep}  \\ \cline{2-4}
   &Representation learning strategy &Medical image segmentation  & \cite{karimijafarbigloo2024mmcformer}  
  
                                                                                   \\ 
                                                                                   \bottomrule
\end{tabular}}}
\end{center}
\end{table*}

%% file: balance/balanced_mm_ver2.tex
\section{Balanced Multimodal Learning}
Different modalities are tightly correlated, since they depict the same concept from different views. This property inspires the flourish of multimodal learning, where multiple modalities are integrated and expected to enhance the understanding of related events or objects. However, despite the natural cross-modal correlation, each modality has distinct data sources and forms. For instance, audio data is typically represented as one-dimensional waveforms, while visual data consists of images composed of pixels. On the one hand, this discrepancy endows each modality with different properties, such as convergence speed, and then makes it hard to process and learn all modalities well simultaneously, bringing difficulty to joint multimodal learning. On the other hand, this discrepancy also reflects in the quality of unimodal data. Even though all modalities depict the same concepts, their information related to the target event or objects is in different amounts. For example, considering an audio-visual sample of label \emph{meeting}, the visual data apparently show the visual content of the meeting, which is easy to recognize (see Figure.1c). While the corresponding audio data is a noisy street car sound, which is hard to build correlation with the label \emph{meeting}. The information amount of visual modality is clearly more than that of audio modality. Due to the greedy nature of deep neural networks~\cite{wu2022characterizing}, multimodal model would tend to rely on just high-quality modality with sufficient target-related information, while under-fitting the other modalities. To address these challenges and improve the efficacy of multimodal models, recent studies have focused on strategies to balance the differences between modalities and enhance the overall performance of the model.

\begin{figure}[t]
\centering
\includegraphics[width=0.8\linewidth]{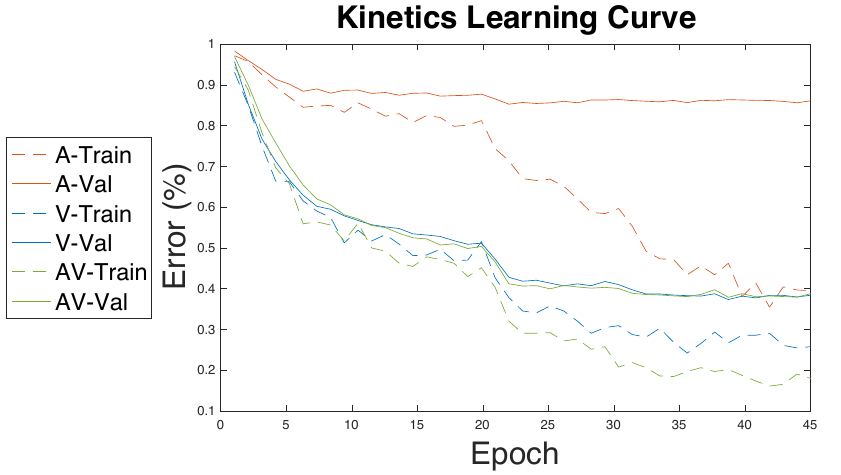}
\caption{The learning curves (error-rate) of audio model (A), video model (V), and the naive joint audio-video (AV) model on the Kinetics dataset. Solid lines plot validation error while dashed lines show train error. Figure is from~\cite{wang2020makes}}
\label{fig:discrepancy}
\end{figure}

\subsection{Property-discrepancy method}
In most multimodal models, different modalities are commonly trained jointly with uniform learning objectives. While this joint training strategy has been generally proven to be effective in practice, there are still potential risk brought by the discrepancy in data of multiple modalities. The discrepancy of multimodal data in sources and forms result in that each modality has its own learning property. For example, as shown in~\autoref{fig:discrepancy}, in the action recognition dataset Kinetics~\cite{kay2017kinetics}, which contains audio and visual modalities, the audio data tends to require fewer training steps than the vision data before the model begins to be overfitting~\cite{wang2020deep, xiao2020audiovisual}. The widely-used joint training multimodal framework often neglects this inherent learning property discrepancy of unimodal data, which can potentially lead to negative impacts on the model's performance. More importantly, based on recent study~\cite{wang2020makes}, the jointly trained multimodal model even could be outperformed by the best unimodal counterpart. Thus, it is essential to consider the discrepancy in learning properties across each modality to improve the performance of the model. To address this issue and balance the discrepancy in learning properties, several studies have been proposed from different perspectives, including learning-objective-based method, optimization-based method and architecture-based method.

\subsubsection{Learning-objective-based method}
Since jointly trained multimodal frameworks often only have a uniform loss, which integrates the information of all modalities, the learning of each modality is hard to observe or control. The learning-objective-based methods add unimodal loss and make it possible to targetedly adjust unimodal training:
\begin{equation}
    \mathcal{L}=\alpha_{mm} \mathcal{L}_{mm} + \alpha_{i} \sum^n_i \mathcal{L}_{i},
\end{equation}
where $\alpha_{mm}$ and $\alpha_{i}$ is the weight of loss function. $n$ is the number of modalities. Since the loss weight determines the influence of the corresponding loss function in parameter update, its selection is essential for controlling unimodal training. Wang \emph{et al.}~\cite{wang2020makes} suggested to determine the optimal loss weights by measuring the proposed overfitting-to-generalization ratio, by which it could comprehensively consider and balance the discrepancy in convergence speed of each modality.

\subsubsection{Optimization-based method}
Different from the learning-objective-based methods that directly introduce unimodal loss, optimization-based method pays attention to the back-propagation stage. Considering the discrepancy in learning property, optimization-based methods propose to control the unimodal parameter updating by dynamically balancing the learning rates for various modalities~\cite{sun2021learning}:
\begin{equation}
    \theta^i_{t+1} = \theta^i_{t} - \lambda_i g^i_{t},
\end{equation}
where $\lambda_i$ is the learning rate of the parameter of modality $i$ and $g^i_{t}$ is the corresponding gradient. In this way, it is possible to conduct modality-specific optimization for each modality. Specifically, Sun \emph{et al.}~~\cite{sun2021learning} proposed to balance the optimization of different modalities by assigning a lower learning rate to the modality that is closer to convergence.

\subsubsection{Architecture-based method}
Besides the above architecture-agnostic methods which focus on learning objective and optimization, methods considering the discrepancy in learning property problem from the perspective of model architecture are also proposed. At the fusion stage, Xiao \emph{et al.}~\cite{xiao2020audiovisual} proposed to randomly drop the network pathway of the modality with faster convergence speed altogether with a certain probability. In this way, the training of the corresponding modality is slowed down and its learning dynamics are ensured to be more compatible with other unimodal counterparts. Besides the late fusion architecture, studies that utilize low-level cross-modal fusion to enhance multimodal joint learning are also proposed. Zhou \emph{et al}.~\cite{zhou2021joint} utilized central connections to link low-level features in both video and audio streams, capturing spatial (for video frames) and temporal information, thereby achieving high-level semantic representations and improving joint multimodal learning.

\begin{figure}[t]
\centering
\subfigure[Audio.]{\includegraphics[width=0.8\linewidth]{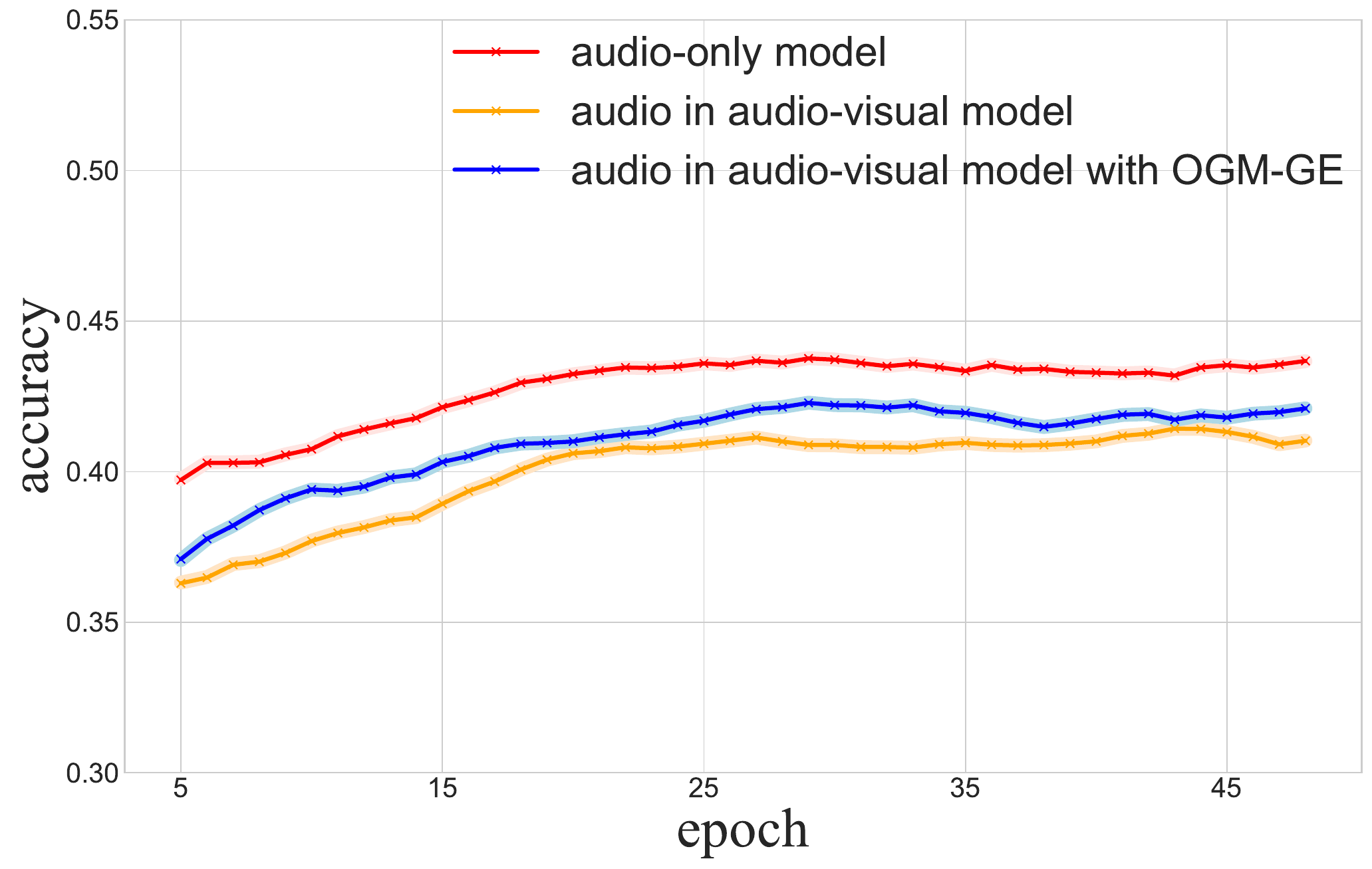}}
\subfigure[Visual.]{\includegraphics[width=0.8\linewidth]{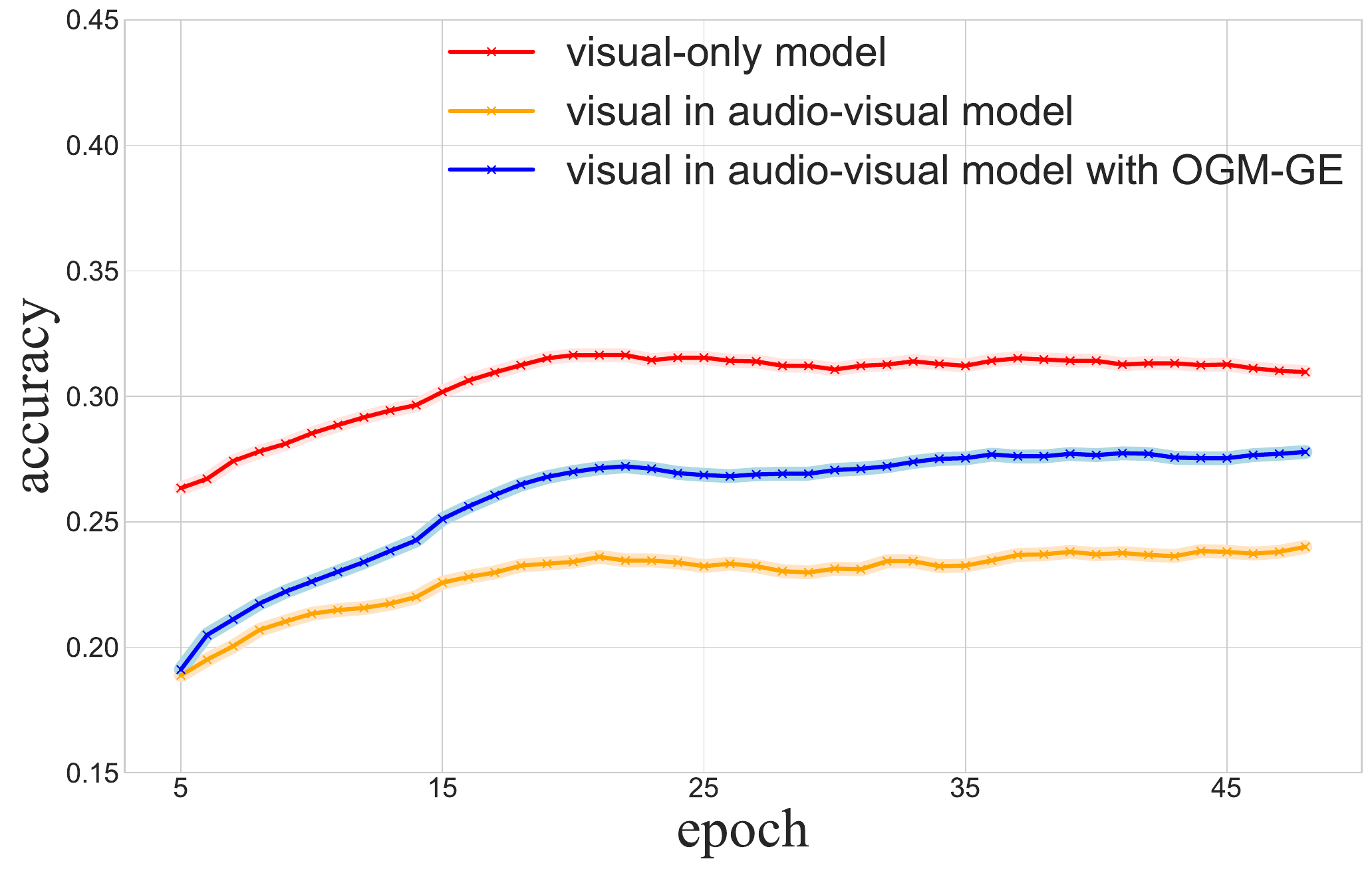}}
\caption{Performance of the uni-modal models, joint-trained multimodal model, and multimodal model with OGM-GE~\cite{peng2022balanced} on the validation set of the VGGSound dataset. Figures are from~\cite{peng2022balanced}.}
\label{fig:imbalance}
\end{figure}

\input{balance/table}

\subsection{Quality-discrepancy methods}
In addition to the issue of discrepancy in unimodal learning property, even though all modalities convey the same concepts, they contain varying amounts of information related to the target event or objects, resulting a discrepancy in unimodal data quality. Given the inherently greedy nature of deep neural networks, recent studies have further uncovered that multimodal models tend to rely heavily on a single modality with high quality that provides sufficient target-related information, while neglecting the others, leading to their under-optimized~\cite{peng2022balanced,huang2022modality}. As shown in Fig.~\ref{fig:imbalance}, Peng \emph{et al.}~\cite{peng2022balanced, wei2024fly} empirically verified that the quality of unimodal representations in the multimodal model is worse than that in the corresponding unimodal models, despite the overall superior performance of the multimodal models. Moreover, this under-optimization of each modality is imbalanced, and the modality with high quality is less under-optimized. They claim that this imbalanced problem is caused by the model preference for the high-quality modality with better performance (i.e., the dominant modality). For instance, the audio modality in the curated sound-oriented dataset, VGGSound~\cite{chen2020vggsound}, is less under-optimized compared to the vision modality. Theoretical analysis conducted by Huang \emph{et al.}~~\cite{huang2022modality} has further confirmed that for the jointly trained multimodal late-fusion framework, the encoder networks can only learn a subset of modalities, and other modalities may fail to be discovered well. This finding is consistent with empirical observations, indicating that the imbalanced preference for different modalities issue needs to be addressed in multimodal learning. To tackle this problem brought by discrepancies in unimodal data quality, a series of studies have been proposed, encompassing learning-objective-based method, optimization-based method, architecture-based method and data-augmentation-based method.

\subsubsection{Learning-objective-based method}
Considering the preference of multimodal model for modality with high quality, several methods which introduce additional learning objectives beyond normal joint learning loss function are proposed. These methods tend to utilize additional loss to break the modality preference of multimodal model:
\begin{equation}
    \mathcal{L}=\mathcal{L}_{mm} + \mathcal{L}_{uni},
\end{equation}
where $\mathcal{L}_{mm}$ is the multimodal loss function and $\mathcal{L}_{uni}$ is a specifically designed loss function that improves unimodal learning, especially the learning of modality with lower quality. Concretely, Yang \emph{et al.}~\cite{yang2022mcl} endowed the multimodal contrastive learning loss to constrain the audio and visual modality to be as close as possible in the semantic space, then help to make full use of all modalities. Also paying attention to the multimodal semantic space, Ma \emph{et al}.~\cite{ma2023calibrating} proposed a novel regularization technique Calibrating Multimodal Learning, to enforce the consistency between prediction confidence and the number of modalities, avoiding the issue that multimodal classification models tend to over-rely on specific modalities. In addition to these methods in semantic space, some methods introduce the learning objective of knowledge distillation. Du \emph{et al.}~\cite{du2023uni} proposed to distill the pre-trained uni-modal features to the corresponding parts of multimodal late-fusion models, to avoid the learning preference for one modality. Differently, Liu \emph{et al.}~\cite{liu2023multimodal} designed a self-distillation training strategy, which can automatically teach less-optimized modality from better-optimized modality. Besides these common multimodal classification task, Xu \emph{et al}~\cite{xu2023mmcosine} focused on the audio-visual fine-grained classification task, which has a higher demand for distinguishable feature distribution. They proposed the multimodal Cosine loss, MMCosine, which performs a modality-wise $L2$ normalization to features and weights towards balanced and better multimodal fine-grained learning. Beyond the generalization performance, Yang \emph{et al}~\cite{yang2024Quantifying} put attention on the multi-modal robustness. They emphasized that multi-modal models are often vulnerable to attacks on the specific modality, and introduced margin-based regularization to ensure multi-modal robustness.

\subsubsection{Optimization-based method}
Optimization-based methods focus on the back-propagation stage. These methods respectively pay attention to the magnitude and direction of unimodal gradient, to improve the learning of modality with lower quality. To slow down the learning of preferred modality with high quality and accordingly offer more training efforts to other modalities, Peng and Wei \emph{et al.}~\cite{peng2022balanced, wei2024fly} proposed an on-the-fly gradient modulation strategy, which dynamically monitors the contribution difference of various modalities to the final prediction during training and mitigates the gradient magnitude of dominant modality to focus more on other modalities. Inspired by this gradient magnitude modulation idea, Sun \emph{et al.}~\cite{sun2023graph} extended it into the graph-based models of rumor detection task, adaptively adjusting the magnitude of gradients to alleviate the preference for high-quality modality. Similarly, Fu \emph{et al.}~\cite{fu2023multimodal} extended the idea of gradient magnitude modulation into audio-visual video parsing task. They further proposed a modality-separated decision unit to well evaluate the unimodal prediction in the more complex audio-visual video parsing framework with more cross-modal interaction. Besides the gradient magnitude, the direction of unimodal gradient is also explored. Fan~\emph{et al.}~\cite{fan2023pmr} suggested that the slow-learning modality’s updating direction is severely disturbed by the preferred high-quality modality, making it hard to its training. Subsequently, they utilize prototypes, which are the centroids of each class within the representation space, to refine the updating direction, aiming to improve unimodal performance.

\subsubsection{Architecture-based method}
Architecture-based methods aim to balance the training of modalities with discrepancy in quality by designing better unimodal representation learning modules. He~\emph{et al.}~\cite{he2022multimodal} designed a Multimodal Temporal Attention module, which takes into account the temporal influences of all modalities on each individual unimodal branch. This module facilitates the interaction among unimodal branches and achieves adaptive inter-modal balance. Su \emph{et al.}~\cite{su2023utilizing} focused on the generative scenarios, and utilized the coordinated feature space, named Coordinated Knowledge Mining, to improve the under-optimized modalities with the guidance of preferred high-quality modalities. Lin~\emph{et al.}~\cite{lin2023variational} proposed a novel Variational Feature Fusion Module, which regards fusion features as random variables and obtains more balanced segmentation performance for different categories and modalities.

\subsubsection{Data-augmentation-based method}
Data-augmentation-based methods are with the idea of enhancing low-quality modality at the data input stage. The basic intuition is that solely training of unimodal data could avoid the influence of other modalities. There are two kinds of strategies, introducing additional individual training for less learnt low-quality modality or removing the data of better learnt high-quality modality. For the former idea, Wu \emph{et al.}~\cite{wu2022characterizing} firstly measured the relative speed at which the model learns from one modality compared to other modalities, and then introduced the re-balancing steps at which update the underutilized uni-modal branches intentionally in order to accelerate the model to learn from its input modality. For the latter one, Zhou~\emph{et al.}~\cite{zhou2023adaptive} suggested adaptively masking the data input of well-learnt modality, accordingly encouraging the model to fit other modalities better. Most imbalanced multi-modal learning methods ignore the fine-grained modality discrepancy, where the dominant modality could vary across different samples under realistic scenarios. Wei~\emph{et al.}~\cite{wei2024enhancing} introduced a Shapley-based sample-level modality valuation metric, to observe the contribution of each modality during prediction for each sample, and then dynamically re-sample low-contributing modalities during training. This method achieves fine-grained multi-modal cooperation.

\subsection{Discussion}
Overall, balanced multimodal learning methods focus on the discrepancy in learning property or data quality among different modalities, which is caused by the heterogeneity of multimodal data. These methods propose solutions from different perspectives, including learning objective, optimization, architecture, and data augmentation. The balanced multimodal learning is still a sunrise field, with plenty of under-explored directions. For example, it is promising to explore the cooperation between modalities with theoretical guidance. Moreover, the current methods mainly limit to the typical multimodal task, mostly discriminative tasks and a few generative tasks. In fact, beyond these, the Multimodal Large Language Model, which also needs to jointly integrate different modalities, could also suffer from this imbalance problem. It is expected to extend current studies or design new solutions in Multimodal Large Language Model settings.

%% file: balance/table.tex
\begin{table*}[htbp]
\centering
\caption{A summary of representative balanced multimodal learning studies.}
\resizebox{0.8\textwidth}{!}{
\setlength{\tabcolsep}{1mm}{
\begin{tabular}{c|c|c|c}
\toprule
                   \multicolumn{2}{c|}{\textbf{Category}}  & \textbf{Application} &   \textbf{Reference}                         \\ \midrule[0.7pt]
 \multirow{4}{*}{\text{\shortstack{Property \\discrepancy}} } & \multirow{2}{*}{Optimization-based} & Action classification               &  \cite{wang2020makes}                          \\\cline{3-4}
             &             & Sentiment analysis       &     \cite{sun2021learning}                   \\\cline{2-2} \cline{3-4}
  & \multirow{2}{*}{Architecture-based}       & Action classification   &    \cite{xiao2020audiovisual}                        \\\cline{3-4}
  &        & Deepfake detection     &    \cite{zhou2021joint}                                     \\ \midrule
 \multirow{12}{*}{\text{\shortstack{Quality\\ discrepancy}}}      & \multirow{5}{*}{Learning-objective-based} & Violence Detection      &  \cite{yang2022mcl}                                        \\\cline{3-4}
                  &  & Action classification   &   \cite{du2023uni}                             \\\cline{3-4}
     &  & Product understanding    &  \cite{liu2023multimodal}                                 \\\cline{3-4}
   &  & Audio-visual fine-grained classification & \cite{xu2023mmcosine}                      \\\cline{3-4}
      &     & Action classification  &  \cite{yang2024Quantifying}             \\\cline{2-2}\cline{3-4}
 &    \multirow{3}{*}{Optimization-based}           & Action classification, sentiment analysis   & \cite{peng2022balanced,fan2023pmr}     \\\cline{3-4}
      &            & Rumor detection           &   \cite{sun2023graph}                     \\\cline{3-4}
      &            & Audio-visual video parsing  &  \cite{fu2023multimodal}                               \\\cline{2-2}\cline{3-4}
        &   \multirow{3}{*}{Architecture-based}       & Sentiment analysis   &   \cite{he2022multimodal}                             \\\cline{3-4}
      &        & Multimodal Conditional Image Synthesis  & \cite{su2023utilizing}                     \\\cline{3-4}
     &        & RGB-T Semantic Segmentation   &   \cite{lin2023variational}                          \\\cline{2-2} \cline{3-4}
     & \multirow{3}{*}{Data-augmentation-based} & Muli-view classification, gesture classification & \cite{wu2022characterizing}              \\\cline{3-4}
     &  & Emotion recognition       &      \cite{zhou2023adaptive}                             \\ \cline{3-4}
     &  & Action classification, emotion recognition &  \cite{wei2024enhancing}                                       \\ \bottomrule
\end{tabular}}}
\end{table*}

%% file: dynamic/dynamic.tex
\section{Dynamic Multimodal Fusion}
\begin{figure*}[!t]
    \centering
    \includegraphics[width=0.90\textwidth]{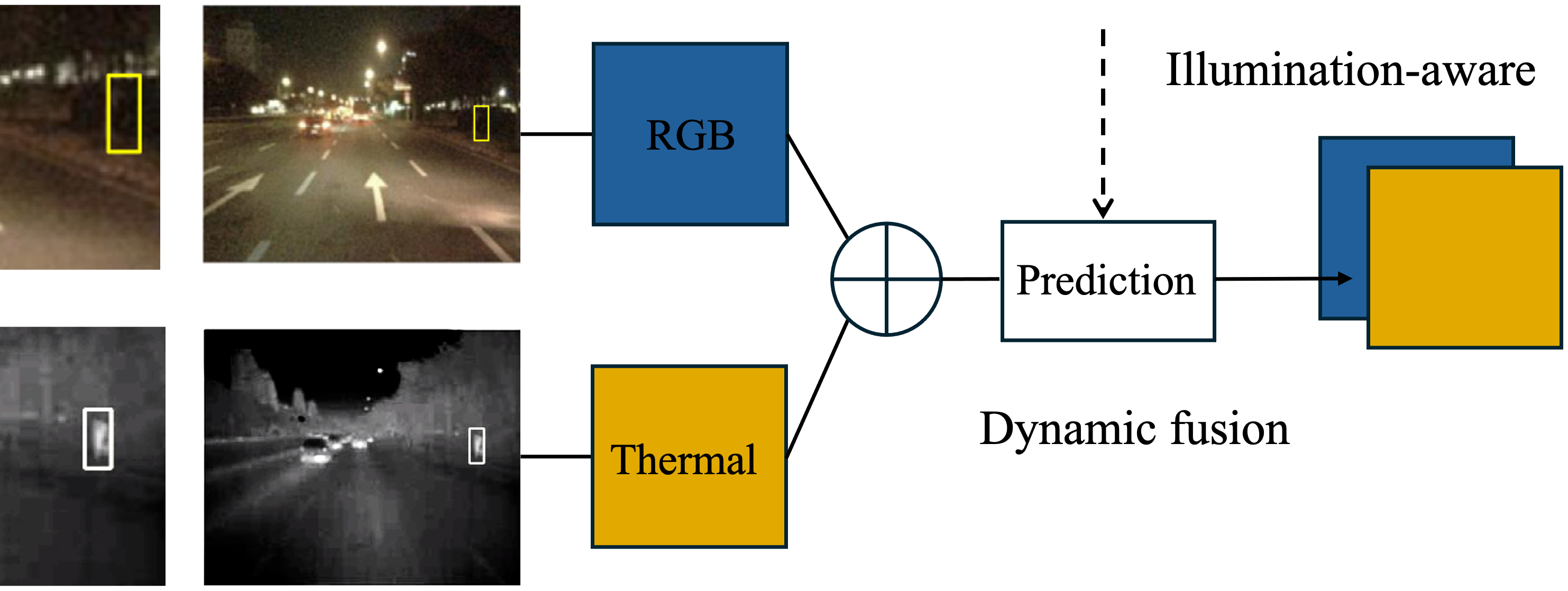}
    \caption{\textbf{Illustrations of dynamic fusion.}} 
    \label{fig:dynamic}
\end{figure*}
Due to unexpected environmental factors and sensor issues, some modalities may suffer from poor reliability and lack of task-specific information. Besides, the quality of different modalities are dynamically varying from scenarios. We borrow a case from~\cite{zhou2020improving}, as shown in Fig.~\ref{fig:dynamic}, thermal modality tends to provide more task-specific information compared to RGB modality during nighttime, whereas the situation is reversed during the daytime. This phenomenon motivates a new multimodal learning paradigm, i.e., dynamical multimodal fusion, which aims to adapt to the dynamically changing quality of multimodal data and selectively integrate task-specific information. In this section, we focus on the challenges of dynamical multimodal fusion and categorize current advances in this literature into three main lines including heuristic, attention-based and uncertainty-aware dynamic fusion.

\input{dynamic/taxon-table}

\subsection{Heuristic dynamic fusion}
Toward achieving robust multimodal fusion, previous works heuristically devise dynamic fusion strategies. These dynamic fusion methods are derived from human experience and knowledge (e.g., thermal modality is more reliable than RGB in nighttime) about the task at hand and application scenarios. As a concrete example, illumination conditions can serve as a criterion to introduce dynamic fusion. In around-the-clock applications, previous work identifies that the quality of RGB and thermal modalities are temporally varying in multispectral pedestrian detection task~\cite{zhou2020improving}. While the RGB modality tends to contain more useful information compared to the thermal modality under normal illumination conditions, this relationship can be reversed under low light or nighttime conditions where the thermal modality becomes more reliable than RGB. Based on this observation, the authors proposed to utilize an illumination aware fusion module that fuse features from the two modalities adaptively. Similarly, to enable adaptation to varying illumination conditions, Guan \emph{et al}.~\cite{guan2019fusion} proposed an illumination-aware weighting mechanism that evaluates the lighting of each input scene. These illumination weights are utilized to dynamically integrate the outputs of sub-networks specialized for day and night detection and segmentation. The proposed method can further be applied to pedestrian detection in security surveillance and autonomous driving systems.

Besides environmental factors like illumination, the criterion of dynamic fusion can be devised according to knowledge about the properties of neural networks. When vision models are involved, activation level of the feature maps and the scaling factors in batch-norm layers can indicate input informativeness. Inspired by this, Li \emph{et al}.~\cite{li2018densefuse} proposed an encoder-decoder style network with dynamic fusion strategy. In their implementation, the l1-norm of feature maps is calculated as channel-wise importance, which is then normalized to form a weighting vector that selectively emphasizes informative features from each modality. Wang \emph{et al}.~\cite{wang2020deep} proposed Channel Exchanging Networks (CEN) that dynamically exchange information between modality sub-networks based on channel importance evaluated by the scaling factors of batch-normalize layers. By performing channel exchange within modality-specific regions and sharing convolutional filters, CEN balances unimodal feature learning and multimodal fusion dynamically.

Some relevant works also utilize task-specific modules to achieve dynamic fusion. To highlight a few, Panda \emph{et al}.~\cite{panda2021adamml} introduced AdaMML, an adaptive multimodal learning in efficient video recognition. By dynamically selecting optimal input modalities per video segment, this work contributes to data-dependent multimodal selection for efficient video recognition. In this work, a lightweight policy network is devised to dynamically select optimal input modalities for each video segment. AdaMML utilizes Gumbel-Softmax sampling for differentiable training of discrete decisions. Similarly, Xue \emph{et al}.~\cite{xue2023dynamic} proposed DynMM, a novel approach for dynamic multimodal fusion. It introduces modality-level and fusion-level decisions, allowing adaptive selection of input modalities and fusion operations based on data characteristics.

\subsection{Attention-based dynamic fusion}

The key challenge of dynamic fusion is to design dynamic mechanism to learn a reasonable fusion criterion. To achieve this, another line of methods typically introduce attention mechanism to dynamic fusing multimodal information. In this subsection, we organize various dynamic fusion methods according to different types of involved attention mechanism, including self-attention, channel-attention, spatial-attention and transformer.

Self-attention models dependency within the input sequence by allowing each element to interact with others. In multimodal emotion recognition (involving textual and audio modalities), audio modality tend to convey more task-specific information than text, such as intonation. However, due to the potential influence of background noise, the quality of these two modalities may vary significantly from different samples. Sun \emph{et al}. introduced a novel Multimodal Cross and Self-Attention Network (MCSAN) ~\cite{sun2021multimodal} to dynamically emphasize information from both the linguistic content and acoustic information in speech emotion recognition. Similarly, Hazarika \emph{et al}. devised a self-attention based multimodal fusion strategy to adapt to the varying quality of textual and audio modalities in emotion recognition task~\cite{hazarika2018self}. By weighing modalities at the feature level, the proposed method provides extra stability against noisy inputs.

Usually applied in vision tasks, channel attention dynamically evaluates the importance of different channels to achieve better performance. To adaptively emphasize important multimodal features while suppressing less relevant ones, Hamid \emph{et al}.~\cite{joze2020mmtm} introduced a Multimodal Transfer Module (MMTM) to achieve better overall performance. MMTM is designed for intermediate fusion and can be inserted at different levels of the feature hierarchy, allowing recalibration of channel-wise features of low-quality modality with knowledge from others. A squeeze-then-excitation operation is implemented to generate a joint representation and excitation weights, enabling adaptive feature recalibration.

The motivation of spatial attention is to assess the importance of different locality in the feature map from the channel dimension. Cao \emph{et al}. introduced a novel lightweight fusion module called Channel Switching and Spatial Attention (CSSA) for multimodal object detection~\cite{cao2023multimodal}. The proposed CSSA module efficiently fuses inputs from different modalities using spatial attention. In their implementation, spatial attention is enhanced using max and average pooling to retain unique features while effectively fusing information.

As one of the most advanced variants of self-attention, transformers (based on multi-head self-attention) are frequently used in recent multimodal fusion methods to enhance the expressive power of the model. Nagrani \emph{et al}.~ proposed Fusion Bottlenecks, a new transformer architecture that restricts cross-modal interactions to bottleneck tokens and mid-late fusion layers only~\cite{nagrani2021attention}. Fusion Bottlenecks force the model to share only necessary information between modalities, avoiding expensive full pairwise attention. Key to the approach is the insight that full pairwise attention between all modalities' tokens is often redundant, and condensing interactions through fusion bottlenecks provides an efficient alternative. Girdhar \emph{et al}. leveraged a shared Transformer architecture with shared parameters to handle diverse vision modalities including images, videos, and single-view 3D data. The key idea is to fuse information from different modalities into a shared embedding representation by employing a self-attention mechanism to model cross-modality relationship both temporally and spatially~\cite{girdhar2022omnivore}. Wang \emph{et al}. identified that existing Transformer variants might dilute unimodal modal attention weights on multimodal tasks, leading to sub-optimal overall performance. To overcome this, they propose to dynamically identifies uninformative tokens and replace them with aggregated features from other modalities for effective fusion~\cite{wang2022multimodal}. Unlike prior works that mainly focus on enhancing reliability of multimodal fusion during training time, Yang \emph{et al}.~\cite{yang2023test} proposed to utilizing attention to fully handle corrupted modalities in test time. A confidence-aware loss is devised to enhance credible predictions and mitigate potential influence of noise.

\subsection{Uncertainty-aware dynamic fusion}

Contrast to heuristic dynamic multimodal learning methods based on intuitive assumptions which might not always hold in practice, uncertainty-based multimodal fusion emerges as a more general and principled way to achieve reliable fusion very recently, which is usually built on solid foundations, e.g., probability distribution or information theory.

\textbf{Subjective Logic} is a commonly used uncertainty estimation method in classification task and a framework that relates the parameters of a Dirichlet distribution to a belief distribution. The Dirichlet distribution can be considered as the conjugate prior distribution of a categorical distribution. SL provides a powerful tool for reasoning under uncertainty and making decisions based on incomplete or uncertain information. Motivated by SL, a Trustworthy Multimodal Classification (TMC)~\cite{han2021trusted}  method was proposed recently based on subjective logic and evidence theory. It provides a new paradigm for multimodal classification by dynamically integrating different modalities with Dempster-Shafer (D-S) theory. Taking two modalities scenario as an example, the multimodal belief masses are calculated in the following manner:
\begin{equation}
\begin{aligned}
\mathcal{M}=\mathcal{M}^1\oplus \mathcal{M}^2,
\end{aligned}
\end{equation}
where $\mathcal{M}=\{\{b_k\}_{k=1}^{K},u\}$ is the multimodal brief mass, $\mathcal{M}^1=\{\{b_k^1\}_{k=1}^{K},u^1\}, \mathcal{M}^2=\{\{b_k^2\}_{k=1}^{K},u^2\}$ are unimodal brief masses and $b^1_k, b^2_k$ is the belief mass assigned for $k$-th class according to each modality. $u^1,u^2$ and $u$ are unimodal and joint uncertainty respectively. Thanks to the property of D-S theory, the final prediction of TMC will depend on the modality with low uncertainty. And thus the classification robustness is improved by integrating evidence from each modality. Inspired by TMC, researchers deployed SL-based uncertainty-aware multimodal fusion in many scenarios, including medical image classification~\cite{feng2022trusted}, object detection~\cite{li2022confidence,li2023stabilizing}, semantic segmentation~\cite{chang2022fast}. In whole-slide medical image classification task, by integrating evidences from modalities of multiple scales, they significantly improve the medical classification performance in autoimmune diseases detection and Fibroma classification tasks. Similarly, Chen \emph{et al}.~\cite{chen2022uncertainty} also leverage Dirichlet distribution to model the uncertainty of multimodal representations and then fuse the segmentation and classification representations adaptively. Although TMC exhibits a great performance, Liu \emph{et al}.~\cite{liu2022trusted} identified its limitations, including inability to guarantee an overall decrease in uncertainty during the fusion of diverse views and neglecting of consistency across views. To address these issues, they proposed an improved opinion aggregation framework, ensuring consistency across views and achieving more reliable predictions.

\textbf{Entropy} is the most natural and straightforward way to capture modal-wise uncertainty. Tian \emph{et al}.~\cite{tian2020uno} proposed an entropy-based multimodal fusion method, named Uncertainty-aware Noisy-Or Multimodal Fusion (UNO). Specifically, UNO first introduces several uncertainty measurements, including predictive entropy, mutual information based on Monte Carlo Dropout, as well as deterministic entropy. It then proceeds to combine these uncertainty metrics for fusion by selecting the most conservative (highest uncertainty) measurement and weighting the outputs from each modalities based on the uncertainty. Finally, a simple but effective noisy-or fusion scheme is deployed to fusing decisions from multiple modalities. In multimodal semantic segmentation task, UNO can be used to improve the robustness in the presence of various unknown input degradation. Similarly, Zhang \textit{et al.}~\cite{zhang2024multimodal} proposed to leverage the entropy of prediction as a proxy to gauge the importance of each modality, and then dynamically weight and fuse the unimodal predictions. By introducing such test-time dynamic fusion mechanism, the proposed method can be aware of uncertainty raised upon the varying quality of different modalities, and further enhance the effectiveness of multimodal fusion.

\textbf{Gaussian distributions} are another commonly used way to model the uncertainty. For example, a multivariate normal distribution can be used to model the distribution of multimodal feature. The variance of estimated distribution are capable of representing modality-wise uncertainty. Based on this point, Geng \emph{et al}.~\cite{geng2021uncertainty} proposed a novel unsupervised multimodal learning method, called Dynamic Uncertainty-Aware Network (DUA-Net). Integrating multiple modalities of noisy data in unsupervised settings is a challenging task. Traditional multimodal methods either treat each modality as equally important, or adjust the weights of different modalities to fixed values, which fail to capture the dynamic uncertainty in multimodal data. By integrating the intrinsic information from multiple modalities under the guidance of uncertainty estimates from a generative perspective, DUA-Net obtains noise-free representations. With the help of uncertainty estimates, DUA-Net can weight each modality of a single sample according to its data quality, thus fully utilizing high-quality samples or modalities while mitigating the influence of noisy samples or modalities. The proposed method is shown to be effective in various unsupervised multimodal learning tasks. For audio-visual emotion recognition, Tellamekala \emph{et al}.~\cite{tellamekala2022cold} proposed Calibrated and Ordinal Latent Distributions (COLD) to model uncertainty for each modality. In this work, the variance of high-dimensional latent distributions is learned first for each modality as a measure of uncertainty. In object detection task, previous work~\cite{chen2022multimodal} proposed a probabilistic ensemble technique (ProbEnsemble) to combine the predictions from multiple detectors, each using different input modalities and architectures. In this work, the authors took the form of a Gaussian with a single variance to measure the uncertainty of unimodal predictions. Then, they probabilistically fused outputs from multiple modalities following Bayes' rule. The proposed probabilistic ensemble framework can estimate the likelihood of each detection proposal and then combine them into a more accurate and robust detection result.

\textbf{Normal-inverse Gamma distribution} hierarchically characterizes the uncertainties with parameters of high-order distributions. In the field of multimodal regression, a recent paper~\cite{ma2021trustworthy} presents a new multimodal fusion approach based on a Normal-Inverse Gaussian (NIG) mixture model. This NIG mixture model is similar to the decision-level fusion method, which combines multiple NIG distributions to output the uncertainty of each modality and the overall uncertainty. In this work, the NIG distribution is deployed to model modal-specific epistemic uncertainty (EU) and aleatoric uncertainty (AU). The contribution of each modality to the final prediction can be dynamically adjusted based on its quality and uncertainty, resulting in more reliable and accurate prediction. The proposed method is implemented using expert decision fusion and achieves significant improvements in both accuracy and uncertainty estimation in multimodal regression tasks.

\textbf{Prediction confidence methods} are designed to directly output the confidence. Different from the uncertainty estimation algorithms, confidence calibration methods aim to obtain confidence by directly calibrating the classification results  without modeling the unknown data distributions. For example, the Maximum Class Probability (MCP)~\cite{hendrycks2016baseline} can be regarded as a baseline method to obtain prediction confidence. Although effective in classification, MCP usually leads to over-confidence especially for erroneous prediction. Therefore, the True-Class Probability (TCP)~\cite{corbiere2019addressing} is employed to obtain more reliable classification confidence. Han \emph{et al}.~\cite{han2022multimodal} proposed a multimodal classification method called Multimodal Dynamics that evaluates both feature-level and modality-level informativeness to selectively fuse multiple modalities in a trustworthy manner. Specifically, it uses a sparse gating strategy and true class probability estimation to assess the varying importance of features and modalities for each sample, achieving more reliable fusion when data quality varies across samples and modalities. Specifically, a modality-level gating strategy is employed to perform dynamic fusion as
\begin{equation}
    \mathbf{h}=[conf^1\cdot \mathbf{h}^1,\cdots,conf^M\cdot \mathbf{h}^M],
\end{equation}
where $[\cdot,\cdot]$ denotes concatenation operator, $\mathbf{h}^1,\cdots,\mathbf{h}^M$ and $\mathbf{h}$ are unimodal and multimodal representations respectively. $conf^m$ is the predicted confidence for the $m$-th modality which measures the informativeness. Zhang \emph{et al}.~\cite{zhang2019weakly} proposed a confidence-aware fusion method to selectively integrate features from RGB and thermal modalities for multi-spectral pedestrian detection. It predicts the classification confidence of each modality, then performs feature re-weighting to emphasize cues from more reliable modalities as well as suppress the less useful ones based on the confidence scores.

\textbf{Theoretical justification} has been proposed to shed light upon the advantage of uncertainty-aware dynamic multimodal learning. In multimodal learning theory literature, seminal work shows multimodal learning is provably better than unimodal from the perspective of generalization ability~\cite{huang2021makes}. Similarly, recent work~\cite{zhang2023provable} illustrates that uncertainty-aware dynamic multimodal learning can establish a tighter generalization error bound compared to traditional static methods. Combining with ensemble-like late fusion strategy in classification setting, we can perform uncertainty-aware weighting fusion in decision-level according to the following rule
\begin{equation}
\label{eq:dynamic fusion}
    f(x)=\sum_{m=1}^M w^{m}(x)\cdot f^m(x),
\end{equation}
where $f^m(x)$ denotes unimodal prediction on modality $m$, $w^m$ is uncertainty-aware fusion weights.
Let $\hat{E}(f^m)$ be the unimodal empirical errors of $f^m$ on $D_{\rm train}$. Then with probability at least $1-\delta\ (1>\delta>0)$, it holds that
\begin{equation}
\begin{split}
\text{GError}(f)\leq\underbrace{\sum_{m=1}^M\mathbb{E}(w^m)\mathbb{E}(f^m)}_{\text{unimodal loss}}+
\underbrace{\sum^M_{m=1}Cov(w^m,l^m)}_{\text{uncertainty-awareness}}+\epsilon,
\end{split}
\end{equation}
where the first term on the right hand side denotes weighted unimodal loss, and the second term is the covariance between fusion weight and loss which represents the uncertainty-awareness ability of the fusion weights. Noted that given an effective uncertainty estimator, the covariance should be negative. This theorem demonstrates that uncertainty-aware dynamic fusion enjoys better generalization ability and further implies a principle to design novel dynamic multimodal fusion methods.

\subsection{Discussion}
Dynamic multimodal learning methods focuse on the variance of modality quality with respect to sample, time or space, which widely exists but tends to be neglected. The dynamic fusion methods include heuristic (mostly designed for specific applications), attention-based (typically for representation fusion) and uncertainty-aware (modeling the modality and sample level uncertainty for fusion) strategies. There are great potentials for dynamic multimodal learning. First, the dynamic principle could be considered in SOTA multimodal models, e.g., CLIP. Second, there are lots of dynamic scenarios in real applications (e.g., autonomous driving, medical image fusion), so designing application-specific dynamic fusion strategies is interesting. For example, in multimodal medical images, we can dynamically fuse them at path-level, which may providing better flexibility and interpretability.

%% file: dynamic/taxon-table.tex
\begin{table*}[htbp]
\setlength{\abovecaptionskip}{0.0cm}  
\setlength{\belowcaptionskip}{0.0cm} 
\vspace{-0.0cm}
\begin{center}
\caption{A summary of our taxonomy of dynamic multimodal fusion \label{self-adaptive}.}
\center\resizebox{0.90\textwidth}{!}{
\setlength{\tabcolsep}{1.9mm}
\begin{tabular}{c|c|c|c}
\toprule
\multirow{1}{*}{\textbf{Type}}   &  \multirow{1}{*}{\textbf{Criterion}} & \multirow{1}{*}{\textbf{Application}} &
    \multirow{1}{*}{\textbf{Reference}}
\\
 \midrule 
 \multirow{13}{*}{Heuristic}  &Illumination condition
           &Pedestrian detection &\cite{guan2019fusion}
            \\ \cline{2-4}
  &Illumination condition
           &Pedestrian detection &\cite{zhou2020improving}
            \\ \cline{2-4}
                                                                          &Reinforcement learning
        &Information extraction &\cite{xu2022different} \\ \cline{2-4}
                                                                         &Activity map
        &Image fusion &     \cite{li2018densefuse} \\     \cline{2-4}     & \multirow{2}{*}{\text{Correlation}}
    &Semantic Segmentation & \multirow{2}{*}{\text{\cite{wang2020deep}}} \\ &  &Image-to-image translation &  \\ \cline{2-4}
                                                                         &Policy Network
          &Audio-visual action recognition &      \cite{panda2021adamml}\\ \cline{2-4}
                                                                            &Feature confidence
        &Multi-omics classification &\cite{zheng2023multi} \\ \cline{2-4}
                                                                               &\multirow{3}{*}{Gating Networks}
           &Sentiment analysis &\multirow{3}{*}{\cite{xue2023dynamic}} \\ & &Semantic segmentation &  \\& &Movie genre classification &  \\ \midrule 
                                                                \multirow{8}{*}{Attention-based}    & \multirow{1}{*}{Self-attention}
       &\multirow{1}{*}{Emotion recognition } &         \multirow{1}{*}{\cite{sun2021multimodal}} \\ \cline{2-4}  & Spatial attention&Object detection & \cite{cao2023multimodal}\\ \cline{2-3} \cline{3-4}
                                                                             & \multirow{3}{*}{\text{\shortstack{Channel attention}}}
    &Audio-visual speech enhancement & \multirow{3}{*}{\text{\cite{joze2020mmtm}}} \\  & &Human action recognition & \\  & &Human gesture recognition &
   \\\cline{2-2} \cline{3-4}
                                                                     &\multirow{5}{*}{Transformer}
       &Semantic segmentation &          \cite{nagrani2021attention}\\\cline{3-3} \cline{4-4}
                                                                     &
       &Classification and retrieval &          \cite{girdhar2022omnivore} \\\cline{3-3} \cline{4-4}
                                                                   &
       &\multirow{3}{*}{\shortstack{Image-to-image translation\\RGBD semantic segmentation\\ 3D object detection}} &         \multirow{3}{*}{ \cite{wang2022multimodal}  } \\& & & \\ & & & \\  \midrule
                                                             \multirow{14}{*}{Uncertainty-aware}        &\multirow{2}{*}{Entropy}
       &Classification &      \cite{zhang2024multimodal}\\ \cline{3-4} & &Semantic segmentation &      \cite{tian2020uno} \\\cline{2-4}
                                                                   &Energy score
           &Sense recognition&            \cite{zhang2023provable}\\ \cline{2-4}    &Bayes' rule
           &Object detection&\cite{chen2022multimodal} 
            \\ \cline{2-4}    &\multirow{5}{*}{Subjective Logic}
        &Sense recognition &\multirow{5}{*}{\shortstack{ \cite{han2021trusted, li2022confidence, chang2022fast}\\\cite{feng2022trusted, liu2022trusted}}}\\ & &Image-text classification &  \\  & &Object detection &  \\ & & Medical image classification & \\ & &Semantic segmentation & \\ \cline{2-4}
      &Gaussian distribution 
       &Emotion recognition &
  \cite{geng2021uncertainty,tellamekala2022cold}\\ \cline{2-4}    &\multirow{1}{*}{NIG Distributions}
           &Sentiment recognition & \multirow{1}{*}{\cite{ma2021trustworthy} } \\ \cline{2-4}    &\multirow{2}{*}{Prediction confidence}
           &Multi-omics classfication & \multirow{2}{*}{\cite{han2022multimodal,zhang2019weakly}} \\ & &Multi-spectral pedestrian detection & \\ \cline{2-4}    &\multirow{1}{*}{Gaussian Process}
           &Multi-view classification &\multirow{1}{*}{\cite{jung2022uncertainty} }

                                                                 \\ \bottomrule
\end{tabular}}
\end{center}
\end{table*}